\documentclass[runningheads]{llncs}
\usepackage{graphicx}

\usepackage{tikz}
\usepackage{comment}
\usepackage{amsmath,amssymb} 
\usepackage{color}

\usepackage{dsfont}
\usepackage{siunitx}
\usepackage{multirow}
\usepackage{booktabs} 
\usepackage{standalone}
\usepackage{xspace}
\usepackage{cleveref}

\newcommand{\myparagraph}[1]{\vspace{4pt}\noindent{\bf #1}}

\usepackage[accsupp]{axessibility}  

\begin{document}
\pagestyle{headings}
\mainmatter

\title{Grounding Visual Representations with Texts for Domain Generalization}
\author{
    Seonwoo Min\inst{1} \and
    Nokyung Park \inst{2} \and
    Siwon Kim \inst{3} \and \\
    Seunghyun Park \inst{4} \and
    Jinkyu Kim \inst{2}
}
\authorrunning{Min et al.}

\institute{
LG AI Research, South Korea \and
Computer Science and Engineering, Korea University, South Korea \and
Electrical and Computer Engineering, Seoul National University, South Korea \and
Clova AI Research, NAVER Corp., South Korea \\
Correspondence: \email{jinkyukim@korea.ac.kr}
}
\maketitle

\begin{abstract}

Reducing the representational discrepancy between source and target domains is a key component to maximize the model generalization. In this work, we advocate for leveraging natural language supervision for the domain generalization task. We introduce two modules to ground visual representations with texts containing typical reasoning of humans: (1) {\em Visual and Textual Joint Embedder} and (2) {\em Textual Explanation Generator}. The former learns the image-text joint embedding space where we can ground high-level class-discriminative information into the model. The latter leverages an explainable model and generates explanations justifying the rationale behind its decision. To the best of our knowledge, this is the first work to leverage the vision-and-language cross-modality approach for the domain generalization task. Our experiments with a newly created CUB-DG benchmark dataset demonstrate that cross-modality supervision can be successfully used to ground domain-invariant visual representations and improve the model generalization. Furthermore, in the large-scale DomainBed benchmark, our proposed method achieves state-of-the-art results and ranks 1st in average performance for five multi-domain datasets. The dataset and codes are available at https://github.com/mswzeus/GVRT.
\keywords{Domain generalization, Image Classification, Textual Explanation, Visual-Textual Joint Embedding}
\end{abstract}

\section{Introduction}\label{sec:intro}

Machine learning systems assume that in-samples (training) and out-of-samples (test) are independent and identically distributed -- this assumption, however, rarely holds in real-world scenarios where domain shift often occurs. Various domain generalization (DG) approaches have been introduced to make models generalize well to unseen novel domains. They mainly focus on learning domain-invariant representations so that the model can leverage such invariances during deployment in unseen test domains. In the DG task, samples from target domains are not available during training, thus these approaches are different from domain adaptation (DA), semi-supervised domain adaptation (SSDA), and unsupervised domain generalization (UDA)~\cite{gulrajani2020search}. 

Reducing the discrepancy between source and target domains is a key component to maximize the model generalization. This is often achieved by (i) explicitly matching the feature distribution across domains using a similarity metric to measure the distance between each domain or by (ii) using contrastive loss to map the latent representations of positive pairs close together and those of negative pairs further away in the feature space. Such distance-based approaches need to optimize all pairwise sample distances, thus potentially resulting in models that are susceptible to outliers and unfair to subgroups in the imbalanced data -- where the classes are not represented equally. 

Distribution of training data also limits the networks' understanding of the data. In computer vision models, their opaque reasoning can be simplified to a situation-specific dependence on visible objects in the image. However, instead of learning the true semantics, they often attend to background objects that are salient to the class labels. (e.g. attending to sea for classifying boats). these models will likely behave well in environments similar to those for which it was trained but typically will not generalize well beyond them~\cite{geirhos2018imagenettrained}


\begin{figure}[!t]
    \begin{center}
        \includegraphics[width=\linewidth]{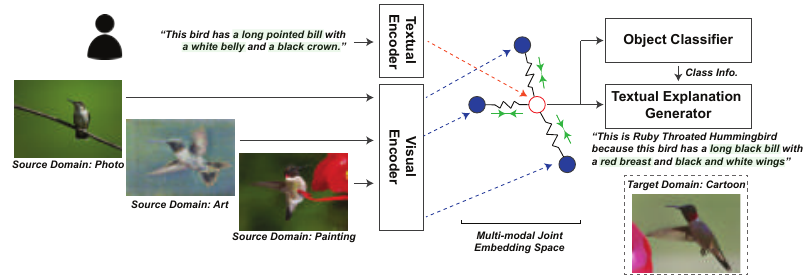}
    \end{center}
    \vspace{-1em}
    \caption{Our model leverages the text modality by (1) {\em Visual and Textual Joint Embedder} and (2) {\em Textual Explanation Generator}. Our model takes advantage of a {\em pivot} embedding (red circle) from a sentence that describes the class discriminative evidence in a natural language, e.g. a white belly or a pointy beak. Our visual encoder is optimized to produce an embedding (blue-filled circles) that aligns well with the corresponding pivot embedding. The latter further trains the model to justify why it made a certain prediction in a natural language.}
    \label{fig:teaser} 
\end{figure}

To address this issue, we propose a novel approach that grounds visual representations with explicit (verbalized) knowledge from humans about typical reasoning on visual cues (Figure~\ref{fig:teaser}). For example, our model learns to understand the user's utterance (``an elephant is a heavy plant-eating mammal with a prehensile trunk, long curved ivory tusks, and large ears.'') and ground it in the trained perceptual primitives. To ground (or internalize) explicit knowledge, we use the following two modules: (1) {\em Visual and Textual Joint Embedder} and (2) {\em Textual Explanation Generator}. The former aligns the perceptual primitives with the (verbalized) thought process of humans by minimizing the distance between the textual and the visual latent representations. The latter leverages the representational power of explainable models. Regardless of image domains, we train the model to consistently verbalize why it made a certain prediction with natural language, e.g. ``This is Ruby Throated Hummingbird because this bird has a long pointed bill with a white belly and a black crown.''

To the best of our knowledge, this is the first work to leverage the vision-and-language cross-modality approach for the DG task. For the empirical evaluations under natural language supervision, we created a new benchmark built upon the Caltech UCSD Birds 200-2011 (CUB) dataset~\cite{WelinderEtal2010}. Our quantitative and qualitative experiment results demonstrate that cross-modality supervision can be successfully used to improve the model representational generalization power as well as to justify its visual predictions. Furthermore, we conducted large-scale experiments on the DomainBed benchmark~\cite{gulrajani2020search}, a popular testbed for DG algorithms. The proposed method achieved state-of-the-art results and ranked ranks 1st in average performance for five multi-domain datasets.

\section{Related Work}\label{sec:relatedwork}

\myparagraph{Domain Generalization.}
Generating domain-invariant representations is the key component in the DG task. Such learned invariances can be leveraged to improve the model generalization to unseen test domains. Of a landmark work, Empirical Risk Minimization (ERM) minimizes the sum of errors across domains, thus matching distributions across different domains~\cite{vapnik1998statistical}. Along this line of work, notable variants have been introduced. DANN~\cite{ganin2016domain} and CDANN~\cite{li2018deep} utilized an adversarial network to minimize unconditional and class-conditional distributional differences across domains, respectively. Such a shared feature space is also optimized by different distance metrics: i.e. maximum mean discrepancy~\cite{li2018domain}, transformed feature distribution distance~\cite{muandet2013domain}, and covariances (CORAL)~\cite{sun2016deep}. Inter-domain mixup techniques~\cite{yan2020improve,xu2020adversarial,wang2020heterogeneous} were introduced to perform ERM on linearly interpolated examples from random pairs across domains. SelfReg~\cite{kim2021selfreg} leveraged the self-supervised learning approaches to address the unstable training, which is often caused by the usage of negative pairs. Furthermore, a recent work theoretically investigated optimal representations for DG~\cite{ruan2022optimal}. They also showed that image-text mapping is a nearly domain-agnostic augmentation that can be leveraged for learning robust representations with self-supervised learning methods.

In this work, we explore the benefit of grounding visual representations by using cross-modality supervision. We introduce two modules for leveraging texts containing the thought process of humans. First, we train a model in the image-text joint embedding space where we can ground high-level class-discriminative information into the model. Second, we adopt an explainable model that can generate explanations justifying the rationale behind its decision.

\myparagraph{Visual and Textual Explanations.}
Explainability and interpretation of deep neural networks have become increasingly important in various machine learning communities~
\cite{gunning2017explainable,kim2020interpretation}. In computer vision, numerous works have explored explaining a target model through visualizations. Early works obtain visual explanations through deconvolutions of layer activations~\cite{zeiler2014visualizing} or synthesizing those that maximize the network output~\cite{zhou2016learning}. Attention-based approaches try to measure how spatial features formally affect the network output~\cite{wang2018deep,wu2018faithful}. They directly extract salient areas of a given image that the network pays the most attention to produce its output. On the other hand, some works emphasize the importance of justifying the model decision in a human-understandable manner, i.e. in natural language. They adopt an encoder-decoder framework which is usually composed of a convolutional neural network (CNN) as the encoder and a long short-term memory (LSTM) caption generator as the decoder~\cite{hendricks2016generating,hendricks2018grounding}. The latter generates textual explanations from the representations produced from the former.


Following this stream of work, we advocate for leveraging the representational power of explainable models for the DG task. Especially, generating textual justifications requires capturing class-discriminative and high-level semantic information. Therefore, we argue that it can help ground domain-invariant visual representations and improve the model generalization.

\section{Method}\label{sec:method}

In this paper, we aim to solve the DG problem: i.e. we train a model on a single or multiple source domains $\{\mathcal{S}_1, \mathcal{S}_2, \dots \}\in\mathcal{S}$ and evaluate it on unseen target domains, $\{\mathcal{T}_1, \mathcal{T}_2, \dots \}\in\mathcal{T}$. Formally, we train a model by minimizing the following data-dependent upper bound on the expected worst-case loss~\cite{sinha2017certifying}:
\begin{equation}
    \underset{\theta}{\mathrm{minimize}}~\underset{\mathcal{T: D(\mathcal{S}, \mathcal{T})\leq\rho}}{\mathrm{sup}}~\mathbb{E}\big[ \mathcal{L}_{\text{task}}(\mathcal{S}; \theta) \big]
\end{equation}
where a dissimilarity $D(\mathcal{S}, \mathcal{T})$ is used to measure the discrepancy between $\mathcal{S}$ and $\mathcal{T}$ with an arbitrary upper bound $\rho$. $\mathcal{L}_\text{task}$ is a task-specific loss function over a model parameter $\theta$ where we use the following cross-entropy loss as we focus on the classification problem:
\begin{equation}
    {\mathcal{L}_\text{task}}(\mathcal{S}; \theta)=\mathcal{L}_\text{task}({\bf{y}}, \hat{{\bf{y}}})=-\sum_i y_i\text{log}(\hat{y}_i)
\end{equation}
where ${\bf{y}}$ is the one-hot vector representing each label's class and $\hat{{\bf{y}}}$ is the softmax distribution produced from the visual feature ${\bf{x}}$. 

\begin{figure*}[!t]
    \begin{center}
        \includegraphics[width=\linewidth]{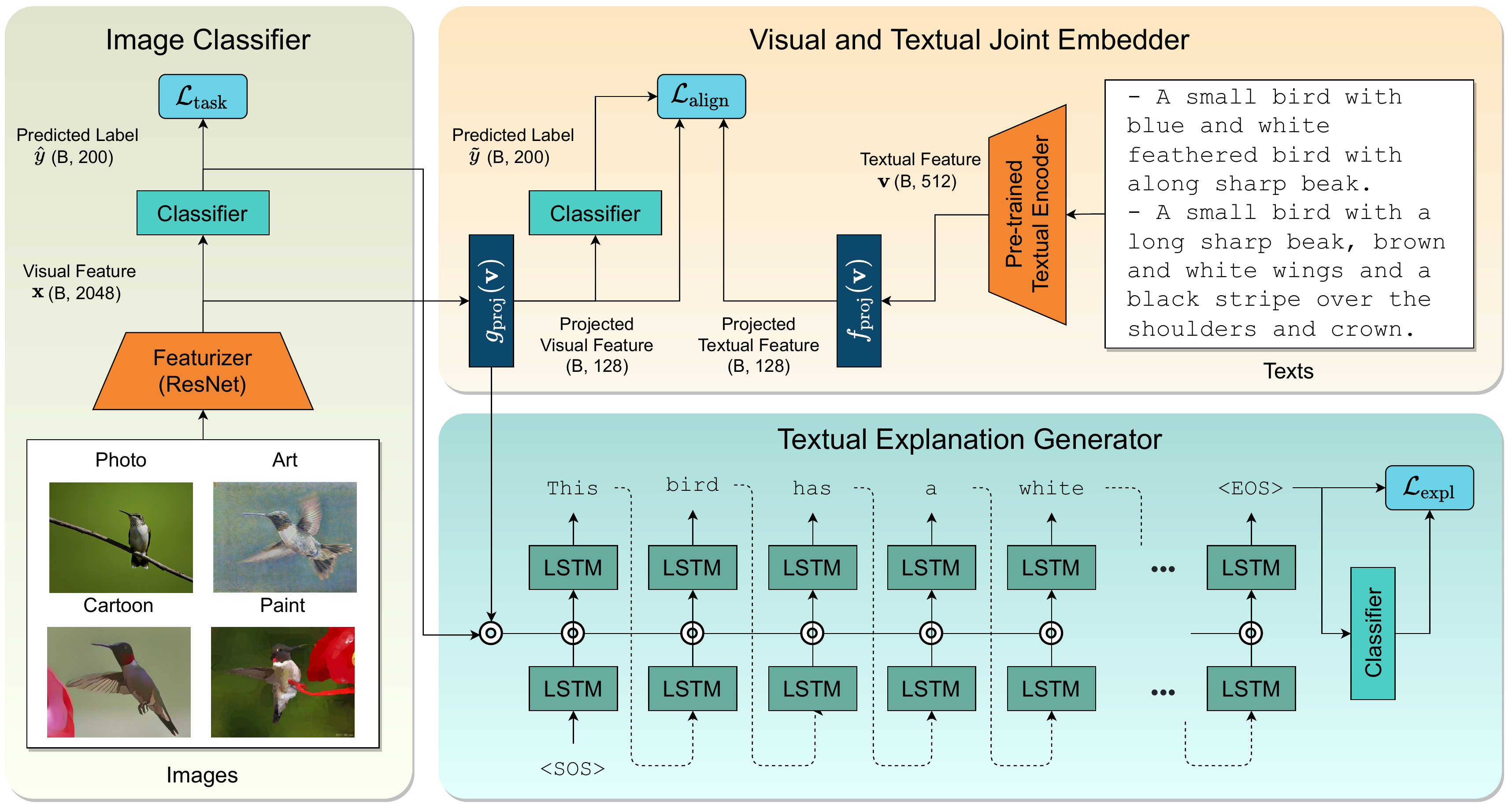}
    \end{center}
    \vspace{-1em}
    \caption{An overview of our proposed model. An image classifier is trained by minimizing the cross-entropy loss $\mathcal{L}_\text{task}$ given images from source domains. Built upon it, our model incorporates two modules for improving DG performance: (i) {\it{Visual and Textual Joint Embedder}}, which produces a joint latent representation that is aligned with textual semantics, and (ii) {\it{Textual Explanation Generator}}, which generates a class-specific sentence detailing how visual evidence is compatible with a system prediction.}
    \label{fig:overview}
\end{figure*}

A key component of the DG task is to address the problem is learning domain-invariant representations that help improve the model generalization. In this work, we advocate for leveraging cross-modality supervision with semantic cues. Specifically, as shown in Figure~\ref{fig:overview}, we use the following two main modules to ground visual representations with texts containing class-discriminative and high-level semantic information. First, {\it{Visual and Textual Joint Embedder}} encourages our visual encoder to produce a latent representation that is aligned with textual semantics in the joint embedding space. Second, {\it{Textual Explanation Generator}} produces a class-discriminative sentence detailing how visual evidence is compatible with a class prediction. Note that both modules are only required during the training phase for grounding the visual encoder. Nevertheless, the latter can be also optionally used during the inference to obtain textual explanation along with a class prediction.

\subsection{Visual and Textual Joint Embedder}
Our Visual and Textual Joint Embedder aims to learn domain-invariant visual representations from textual explanations. We argue that learning from natural language supervision has potential strength over image-only training approaches, especially for the DG task. In contrast to the vulnerability of CNNs against domain shift, the human visual recognition system generalizes well across domains, e.g. even very young children can easily transfer object concepts from picture books to the real world~\cite{ganea2008transfer}. Thus, we advocate for using explicit knowledge from humans and we train a model to better align with the thought process of humans via their textual explanations. It ultimately provides more semantically-rich information compared to standard crowd-sourced labeling for image classification. 

We use a sentence-level textual encoder, which takes a variable-length sentence and yields a fixed-size latent vector ${\bf{v}}$. Given the {\em pivot} textual latent representations ${\bf{v}}$, we optimize the visual encoder to produce representations ${\bf{x}}$ that align well with the corresponding pivot. Thus, our model needs to understand the textual justification from human annotators and to map it into the image-text joint embedding space. We assume that such textual justification will often contain class-discriminative evidence reflecting visual semantic cues, thus our visual encoder can internalize knowledge from natural language supervision.

Specifically, we minimize the following loss function $\mathcal{L}_\text{align}$ based on $l_2$ distance between the projected visual and texture features:
\begin{equation}
    \mathcal{L}_\text{align} = ||f_{\text{proj}}({\bf{v}}) - g_{\text{proj}}({\bf{x}})||_2 -\sum_i y_i\text{log}(\tilde{y}_i)
\end{equation}
where $f_{\text{proj}}$ and $g_{\text{proj}}$ are the projection layers for text and visual feature, respectively. ${\bf{y}}$ is the one-hot vector representing each label's class and $\tilde{y}_i$ is the softmax distribution produced from the projected visual feature $g_{\text{proj}}({\bf{x}})_i$. Note that we use the second cross-entropy term to make the projected visual features more class-discriminative. It prevents collapsing into collapsing solutions, e.g., always projecting them to the same point.

\myparagraph{Pre-trained (Supervised) Textual Encoder (PTE).}
One way to obtain the pivot textual latent representation is via pre-trained language models. These pre-trained encoders can be adopted from off-the-shelf sentence-level textual encoders that are often pre-trained with a large-scale dataset. In this work, we adopt the widely used CLIP (i.e. Contrastive Language-Image Pre-Training) model, which can embed texts and images into the joint representation space~\cite{radford2021learning}. The text encoder of CLIP is a 63$M$-parameter Transformer architecture with \num{12}-layer, \num{512}-wide, and \num{8} attention heads~\cite{vaswani2017attention}. It was jointly trained with a Vision Transformer (ViT)-based image encoder~\cite{dosovitskiy2020image} to predict the pairing of texts and images. In this work, we only used the text encoder of the CLIP-ViT-B-32 model, but other pre-trained language models are also applicable.

\myparagraph{Self-supervised Textual Encoder (STE).}
Another way to obtain the pivot textual latent representation is via self-supervision. Since our textual explanation generator justifies the rationale behind the model in the natural language, we can use it as a self-supervised textual encoder. As we will explain in the next subsection, during the training, we iteratively sample a sentence from an LSTM-based explanation generator to compute its training loss. Therefore, it is an intuitive choice to use its last hidden states as a fixed-size latent vector ${\bf{v}}$.

\subsection{Textual Explanation Generator}
Our textual explanation generator is similar to image captioning models based on an encoder-decoder framework. It contains a two-layer LSTM network that takes high-level features from the visual encoder as input and generates variable-length per-word softmax probabilities. The difference is that it is trained to explain the rationale behind the classifier, reflecting typical visual semantic cues. Since it needs the prediction outputs from the classifier as an input as well, we concatenate the category information with a projected visual feature $g_{\texttt{proj}}({\bf{x}})$. The concatenated vector is then used to update the LSTM network for a textual explanation generation.  

Specifically, the first LSTM layer takes the previously generated output token ${o_{t-1}}$ as input and updates its hidden state, producing an output ${\bf{z}_t}$. This output is then fed into the second LSTM layer along with the concatenated vector of projected visual features and prediction outputs. The second LSTM layer yields the per-word softmax probabilities $p(o_t)$. Further, following \cite{hendricks2016generating}, we use the discriminative sentence generation loss function based on reinforcement learning so that a model learns to generate sentences that are more likely to be class-discriminative. Specifically, we first sample a sentence from the textual explanation generator and we minimize the expectation of the negative reward $-R(\tilde{o})$ over the sampled sentences $\tilde{o}\sim p(o|\mathcal{I}, \mathcal{C})$. The probability distribution $p(o|\mathcal{I}, \mathcal{C})$ is the model's estimated conditional distribution over descriptions $o$ conditioned on the input image $\mathcal{I}$ and the category $\mathcal{C}$. Concretely, for training our textual explanation generator, we minimize the following loss function $\mathcal{L}_\text{expl}$:
\begin{equation}
    \mathcal{L}_\text{expl} = -\sum_{t}\log p(o_{t+1}|o_{0:t}, \mathcal{I}, \mathcal{C}) - \mathbb{E}_{\tilde{o}\sim p(o|\mathcal{I}, \mathcal{C})}\big[ R(\tilde{o}) \big] 
\end{equation}
We use the reward function as $R(\tilde{o})=p(\mathcal{C}|\tilde{o})$, which is the per-class softmax probabilities over the category $\mathcal{C}$ conditioned on the generated sentence $\tilde{o}$. A more class-discriminative sentence receives a higher reward. Using REINFORCE~\cite{williams1992simple} algorithm, we compute the following expected reward gradient as:
\begin{equation}
    \nabla_{\theta}\mathbb{E}_{\tilde{o}\sim p(o|\mathcal{I}, \mathcal{C})}\big[ R(\tilde{o}) \big] = \mathbb{E}_{\tilde{o}\sim p(o|\mathcal{I}, \mathcal{C})}\big[ R(\tilde{o}) \nabla_{\theta} \log p(\tilde{o}) \big]
\end{equation}

\myparagraph{Loss function.}
To summarize, we train our entire model end-to-end by minimizing the following loss function $\mathcal{L}$:
\begin{equation}
    \mathcal{L} = \mathcal{L}_\textnormal{task} + \lambda_\textnormal{align}\mathcal{L}_\textnormal{align} + \lambda_\textnormal{expl}\mathcal{L}_\textnormal{expl} 
\end{equation}
where we use hyperparameters $\lambda_\textnormal{align}$ and $\lambda_\textnormal{expl}$ to control the strengths of each training objective term.

\section{Caltech UCSD Birds - Domain Generalization Extension (CUB-DG) Dataset}

\begin{figure}[!t]
    \begin{center}
        \includegraphics[width=\linewidth]{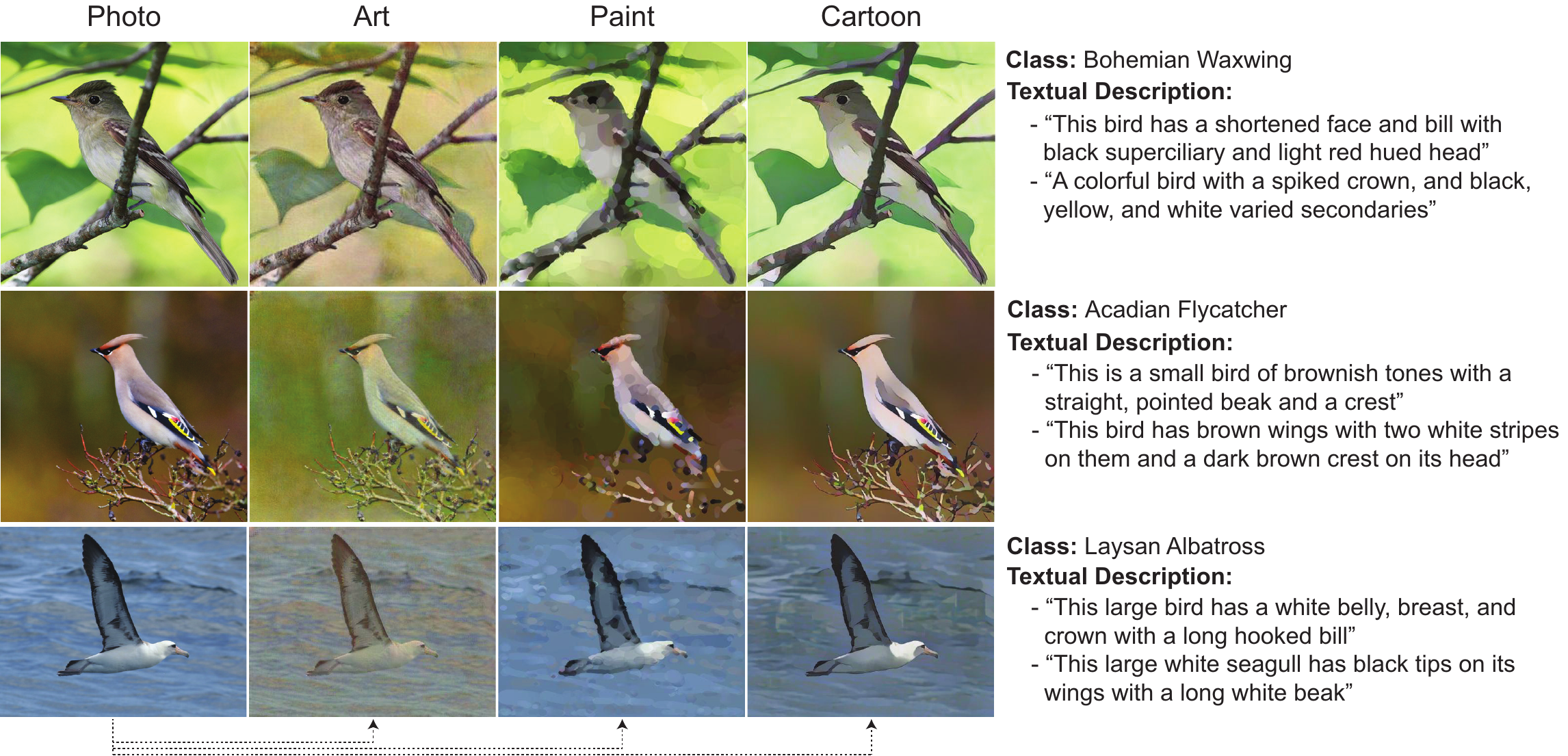}
    \end{center}
    \caption{We create an extended dataset for the DG task based on Caltech UCSD Birds 200-2011 (CUB) dataset. This dataset provides a pair of images and detailed descriptions of the content of the image, e.g. a description ``this is a small bird of brownish tones with a straight, pointed beak and a crest'' for Acadian Flycatcher. On top of this dataset, we applied off-the-shelf style transfer techniques to obtain images from three other domains: Art, Paint, and Cartoon.}
    \label{fig:dataset}
\end{figure}
%
No previous DG benchmarks provide viable natural language supervision. Thus, in order to thoroughly investigate the effectiveness of the cross-modality supervision in the DG task, we have created a new benchmark built upon the CUB dataset~\cite{WelinderEtal2010}. This dataset contains overall 11,788 images for 200 classes of North American bird species. Ten sentences for each of the images have been previously collected~\cite{reed2016learning}, which provides a detailed description of the content of the image, e.g., ``this bird has a long pointed bill with a white belly and a black crown.'' This dataset has been an ideal benchmark for the visual explanation task as sentences are class-specific and class-discriminative. 

\myparagraph{CUB Dataset for Domain Generalization}
Since the CUB dataset is only composed of the Photo domain, we used pre-trained style transfer models to obtain images from three other domains, i.e. Art, Paint, and Cartoon. For the Photo-to-Art translation, we used the CycleGAN~\cite{zhu2017unpaired} \textit{Monet} model which was trained in the absence of paired examples based on adversarial and cycle-consistency losses. For the Photo-to-Paint translation, we used the \textit{Watercolor} neural render model~\cite{zou2021stylized}. It imitates painting creation processes by producing a sequence of strokes. For the Photo-to-Cartoon translation, we used a generative adversarial network model which separately identifies surface, structure, and texture representations of cartoons~\cite{wang2020learning}.

The generated CUB-DG dataset contains 11,768 sets of images and corresponding text descriptions. Each set illustrates the same content of a bird in four different domains. To evaluate DG algorithms in common experimental protocols, we used the following data split procedures (Figure S1). We start from the official split of the CUB dataset where the train-validation and test sets consist of 5,994 and 5,794 samples, respectively. We divide the train-validation set into three groups. For the multi-source DG task, we select a different group from each source domain, so that the different domains do not share the \textit{siblings} of the same image. For the single-source DG task, we use all three groups from a source domain. For both tasks, we evaluate DG algorithms on the test set from unseen target domains. Note that the CUB-DG dataset holds evident domain shifts such that an ERM model trained only on the Photo domain performs well on the same domain (71.2 accuracy; results not shown) but significantly deteriorates on the other domains (Table~\ref{Table:single-domain}).

\section{Experiments}\label{sec:experiment}

{
\setlength{\tabcolsep}{4pt}
\renewcommand{\arraystretch}{1.3} 
\begin{table}[t]
	\begin{center}
	    \caption{Out-of-distribution test accuracies on the CUB-DG benchmark dataset. We compare with 12 DG algorithms in the multi-source DG setting. Note that we use the validation set (from source domains) for the model selection. {\it{Abbr.}} $D$: learning domain-invariant features by matching distributions across different domains, $A$: adversarial learning strategy, $M$: inter-domain mix-up, $T$: learning textual representations. PTE: pre-trained (supervised) textual encoder, STE: self-supervised textual encoder.}
	    \label{Table:dg_performance}
    	\resizebox{.7\linewidth}{!}{%
    	\begin{tabular}{@{}lcccccccccc@{}} 
    	    \toprule
        	\multirow{2}{*}{Model} & \multirow{2}{*}{$D$} & \multirow{2}{*}{$A$} & \multirow{2}{*}{$M$} & \multirow{2}{*}{$T$} & \multicolumn{4}{c}{Target Domain} & \multirow{2}{*}{Avg} \\\cmidrule{6-9}
        	& & & & & Photo & Cartoon & Art & Paint &  \\ 
        	\midrule
            Ours w/ PTE &  \checkmark  &  &  &  \checkmark  & \textbf{74.6} & \textbf{64.2} & \textbf{52.2} & 37.0 & \textbf{57.0} \\
            Ours w/ STE &  \checkmark  &  &  &  \checkmark  & 74.3 & 63.9 & 50.0 & \textbf{38.1} & 56.6 \\
            \midrule
            CORAL~\cite{sun2016deep}  &  \checkmark  &   &   &   & 72.2 & 63.5 & 50.3 & 35.8 & 55.4 \\
            SD~\cite{pezeshki2020gradient}  &   &   &   &   & 71.3 & 62.2 & 50.8 & 34.8 & 54.7 \\
            SagNet~\cite{nam2021reducing} &  \checkmark  &  \checkmark  &  &  & 67.4 & 60.7 & 44.0 & 34.2 & 51.6 \\
            MixStyle~\cite{zhou2020domain}  &   &   &  \checkmark  &   & 59.0 & 56.7 & 50.3 & 35.8 & 50.4 \\
            Mixup~\cite{yan2020improve}  &   &   &  \checkmark  &   & 67.1 & 55.9 & 51.1 & 27.2 & 50.3 \\ 
            DANN~\cite{ganin2016domain}  &  \checkmark  &  \checkmark  &   &   & 67.5 & 57.0 & 42.8 & 30.6 & 49.5 \\
            CDANN~\cite{li2018deep}  &  \checkmark  &  \checkmark  &   &   & 65.3 & 55.2 & 43.2 & 30.5 & 48.6 \\
            VREx~\cite{krueger2020out}  &  \checkmark  &   &   &   & 63.9 & 54.9 & 38.6 & 30.1 & 46.9 \\
            ERM~\cite{vapnik1999overview}  &   &   &   &   & 62.5 & 53.2 & 37.4 & 29.0 & 45.5 \\
            ARM~\cite{zhang2020adaptive}  &   &   &   &   & 62.3 & 51.2 & 38.2 & 28.4 & 45.0 \\
            GroupDRO~\cite{sagawa2019distributionally}  &  \checkmark  &   &   &   & 60.9 & 54.8 & 36.5 & 27.0 & 44.8 \\
            IRM~\cite{arjovsky2019invariant}  &   &   &   &   & 60.6 & 51.6 & 36.5 & 30.3 & 44.8 \\
            \bottomrule
        \end{tabular}}
     \end{center}
\end{table}
}

\myparagraph{Multi-Source Domain Generalization Performance.}
We first look into the multi-source DG task, where a single domain is used as a test domain and the others as training domains in rotation. We compare our model with 11 DG algorithms from DomainBed on our newly created CUB-DG dataset. Compared methods include ERM~\cite{vapnik1999overview}, IRM~\cite{arjovsky2019invariant}, GroupDRO~\cite{sagawa2019distributionally}, Mixup~\cite{yan2020improve}, CORAL~\cite{sun2016deep}, DANN~\cite{ganin2016domain}, CDANN~\cite{li2018deep}, SagNet~\cite{nam2021reducing}, MixStyle~\cite{zhou2020domain}, ARM~\cite{zhang2020adaptive}, VREx~\cite{krueger2020out}, and SD~\cite{pezeshki2020gradient}. We report averaged results across three independent runs. Please refer to the supplementary materials for complete implementation details.

We observe in Table~\ref{Table:dg_performance} that our proposed models outperform the other recent approaches in all test domains (compare the top two rows vs. others), and the average image recognition accuracy is 1.6-12.2\% better than alternatives.  While the performance difference between our model variants is marginal, we also observe that our model with the PTE generally shows better performance than a model with the STE. Therefore, in the following, we focus on analyzing our model with the PTE. Our model can be used together with other approaches (e.g. SD~\cite{pezeshki2020gradient} and SWA~\cite{cha2021domain}) that are not based on matching distributions across domains, which would be worth exploring as future work.

{
\setlength{\tabcolsep}{4pt}
\renewcommand{\arraystretch}{1.3} 
\begin{table*}[t]
	\begin{center}
	    \caption{Out-of-distribution test accuracies in the single-source DG setting where we train our model with a single source domain (rows) and evaluate with other remaining target domains (columns). We compare with SD~\cite{pezeshki2020gradient} and report differences between ours in the last row ($+$ indicates that ours performs better).}
	    \label{Table:single-domain}
    	\resizebox{\linewidth}{!}{%
    	\begin{tabular}{@{}lccccclccccc@{}} \toprule
    	    \multirow{2}{*}{{\bf{SD}}~\cite{pezeshki2020gradient}} & \multicolumn{5}{c}{Target Domain} & \multirow{2}{*}{{\bf{Ours}}} & \multicolumn{5}{c}{Target Domain}\\
    	    & Photo & Cartoon & Art & Paint & Avg. & & Photo & Cartoon & Art & Paint & Avg. \\
    	    \midrule
            Photo & - & 42.4 & 51.3 & 20.4 & 38.0 & Photo & - & 49.1 & 54.2 & 19.5 & 40.9 \\
            Cartoon & 66.9 & - & 29.3 & 34.6 & 43.6 & Cartoon & 69.5 & - & 33.6 & 36.3 & 46.5 \\
            Art & 69.0 & 33.4 & - & 15.7 & 39.4 & Art & 75.6 & 37.9 & - & 16.3 & 43.2 \\
            Paint & 58.0 & 49.9 & 30.0 & - & 46.0 & Paint & 63.7 & 57.3 & 35.6 & - & 52.2 \\ 
    	    \midrule
    	    Avg & 64.6 & 41.9 & 36.9 & 23.6 & 41.7 & Avg & 69.6 & 48.1 & 41.1 & 24.0 & 45.7 \\
    	    & & & & & & & (+5.0\%) & (+6.2\%) & (+4.3\%) & (+0.5\%) & (+4.0\%)\\
            \bottomrule
        \end{tabular}}
     \end{center}
\end{table*}
}

\myparagraph{Single-Source Domain Generalization Performance.}
We also evaluate our model in an extreme case for the DG task, i.e. single-source DG. In this setting, we assume that only a single domain is available during the training. We then evaluate with examples from all the other remaining target domains. In Table~\ref{Table:single-domain}, we compare ours with those of SD~\cite{pezeshki2020gradient}. We show differences between ours and SD in the last row ($+$ indicates that ours performs better) and present them as a heatmap in the Figure S2. Additionally, the full results for comparing our model with six DG algorithms are also available in the Table S1. We excluded algorithms that are inapplicable for the single-source DG setting. We report scores for all source-target combinations, i.e. rows and columns for source and target domains, respectively. The scores are averaged across three independent runs. We observe in Table~\ref{Table:single-domain} that ours outperforms alternatives, where the average accuracy is improved by 4.0\% than SD~\cite{pezeshki2020gradient}.

\myparagraph{Ablation Studies.}
To better understand different aspects of our proposed models, we present results from ablation studies. We vary our base model in several directions and measured the performance on the multi-source DG task. We report averaged results across three independent runs. 

First, we vary the amount of natural language supervision. While we have assumed \textit{Per-Image} texts are available, obtaining them across different domains may not be easy in real life. Therefore, we introduce a more practical scenario where we only use the same single sentence for all the images within each class. Intuitively, it can be understood as \textit{Per-Class} textual definitions. In Table~\ref{Table:ablation_studies} row (A), we observe that even with the \textit{Per-Class} texts, the cross-modality supervision still enables outperforming all the compared DG algorithms in Table~\ref{Table:dg_performance}.

In Table~\ref{Table:ablation_studies} rows (B), we investigate the importance of each module. We observe that removing the Joint Embedder significantly hurts the DG performance. The Explanation Generator plays a complementary role in grounding visual representations by justifying model predictions in natural language. In rows (C), we look into the sensitivity to the hyperparameters $\lambda_\textnormal{align}$ and $\lambda_\textnormal{expl}$. We can see that the former is more crucial in the training of our proposed model. We provide more extensive results as a heatmap in the Figure S3. Furthermore, in Table S2, we compare the impact of embeddings from various PTEs. While we use CLIP as default, different PTEs also successfully produce domain-invariant representations.

{
\setlength{\tabcolsep}{4pt}
\renewcommand{\arraystretch}{1.3} 
\begin{table}[t]
	\begin{center}
	    \caption{Results from ablation studies. We vary our base model in several directions and measured the performance on the multi-source DG task.}
	    \label{Table:ablation_studies}
    	\resizebox{0.9\linewidth}{!}{%
    	\begin{tabular}{@{}lcccccccccc@{}} 
    	    \toprule
        	& \multirow{2}{*}{\shortstack{Available\\Texts}} & \multirow{2}{*}{\shortstack{Joint\\Embedder}} & \multirow{2}{*}{\shortstack{Explanation\\Generator}} & \multirow{2}{*}{$\lambda_\textnormal{align}$} & \multirow{2}{*}{$\lambda_\textnormal{expl}$} & \multicolumn{4}{c}{Target Domain} & \multirow{2}{*}{Avg} \\\cmidrule{7-10}
        	& & & & & & Photo & Cartoon & Art & Paint &  \\ 
        	\midrule
            Base & Per-Image & Yes & Yes & 1.0 & 1.0 & 74.6 & \textbf{64.2} & \textbf{52.2} & \textbf{37.0} & \textbf{57.0} \\
            \midrule
            (A) & Per-Class &  &  &  &  & 	\textbf{74.8} & 63.2 & 51.9 & 36.1 & 56.5 \\
            \midrule
            \multirow{2}{*}{(B)}  &    &  No  &    &   &   &  68.5 & 57.2 & 42.3 & 29.1 & 49.3 \\
              &  &    &  No  &  	&    & 73.7 & 63.8 & 50.2 & 36.5 & 56.1 \\
            \midrule
            \multirow{2}{*}{(C)}  &    &    &    &  0.1  &  1.0  &  73.0 & 63.1 & 50.1 & 33.0 & 54.8 \\
              &    &    &    &  1.0  &  0.1  &  73.1 & 63.8 & 50.3 & 36.4 & 55.9 \\
              &    &    &    &  0.1  &  0.1  &  72.0 & 61.3 & 46.7 & 33.9 & 53.5 \\
            \bottomrule
        \end{tabular}}
     \end{center}
\end{table}
}

\begin{figure*}[!t]
    \begin{center}
        \includegraphics[width=\linewidth]{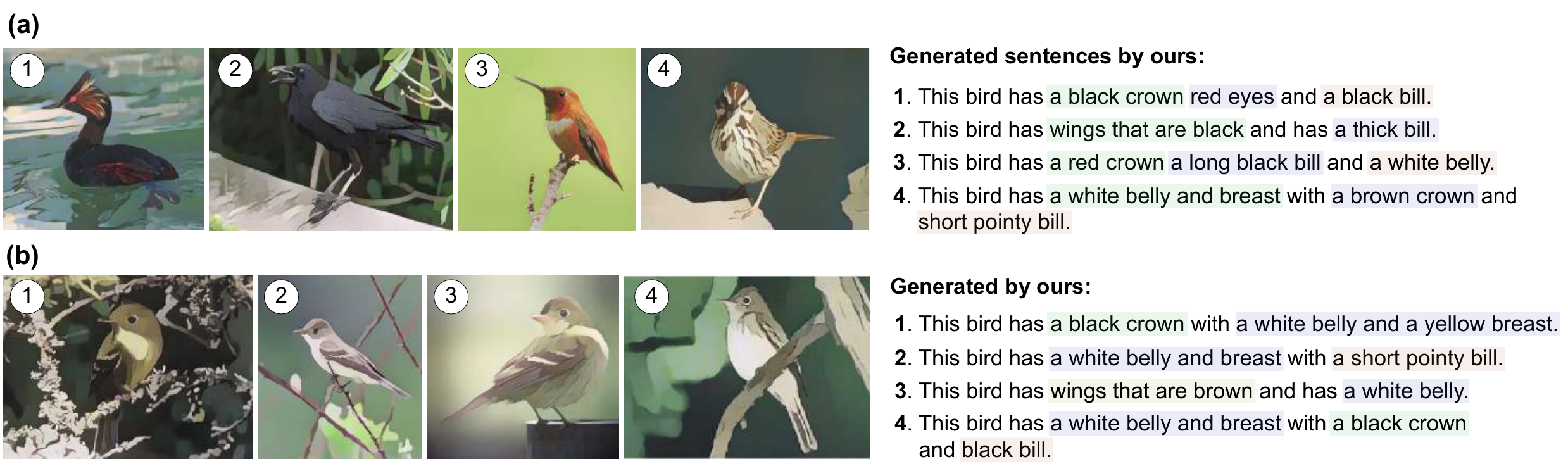}
    \end{center}
    \caption{(a) Textual explanations generated by our model. Our model generates plausible sentences that describe fine details about the class-discriminative attributes. We highlight such attributes with colors. (b) We further compare the generated explanations between different same-class images (i.e. Acadian Flycatcher).}
    \label{fig:generated_captions}
\end{figure*}

{
\setlength{\tabcolsep}{4pt}
\renewcommand{\arraystretch}{1.3} 
\begin{table}[t]
	\begin{center}
	    \caption{We report the quality of the generated textual explanations. We rely on standard metrics: BLEU~\cite{papineni2002bleu}, METEOR~\cite{lavie2005meteor}, CIDEr-D~\cite{vedantam2015cider}, and ROUGE\_L~\cite{lin2004rouge}.}
	    \label{Table:caption_performance}
    	\resizebox{.7\linewidth}{!}{%
    	\begin{tabular}{@{}lcccc@{}} \toprule
    	    Model & BLEU-4 & METEOR & CIDEr-D & ROUGE\_L \\ 
    	    \midrule
        	Ours w/ Joint Embedder & \textbf{48.0} &	\textbf{31.7} & \textbf{40.7} & \textbf{61.8} \\
        	Ours w/o Joint Embedder & 42.9 & 28.0 & 28.4 & 58.1 \\
            \bottomrule
        \end{tabular}}
     \end{center}
\end{table}
}

\myparagraph{Generated Textual Justification Quality.}
Next, we evaluate the quality of our generated textual justification. In Figure~\ref{fig:generated_captions} (a), we provide sample explanations generated by our model. Note that the images shown in the figure are from unseen target domains. The model was trained in the photo, art, and paint domains and tested in the cartoon domain. Qualitatively, our textual explanation generation module accurately describes fine class-discriminative details such as ``red eyes'' or ``white belly and breast.'' These are important and domain-invariant visual cues to determine their classes. For a better understanding, we highlight class discriminative attributes in the generated sentences. 

As shown in Figure~\ref{fig:generated_captions} (b), we further provide generated explanations for different images of Acadian Flycatcher (in the Cartoon domain). As we expected, our model describes fine details of the diverse class-discriminative attributes, which are consistent over different same-class images. This may imply that our network's visual representations are grounded by such consistent cues, which helps in providing the model generalization. For a better understanding, we highlight the same attributes with the same color. 

We further quantitatively evaluate the quality of generated sentences. We use popular metrics: BLEU~\cite{papineni2002bleu}, METEOR~\cite{lavie2005meteor}, CIDEr-D~\cite{vedantam2015cider}, and ROUGE\_L~\cite{lin2004rouge}. These metrics are widely used for the automatic evaluation of image captioning models against ground truth. The scores are averaged across three independent runs. We observe in Table~\ref{Table:caption_performance} that our model with the Visual and Textual Joint Embedder as well as the Textual Explanation Generator obtains higher scores in all metrics than its counterpart.

\begin{figure*}[!t]
    \begin{center}
        \includegraphics[width=\linewidth]{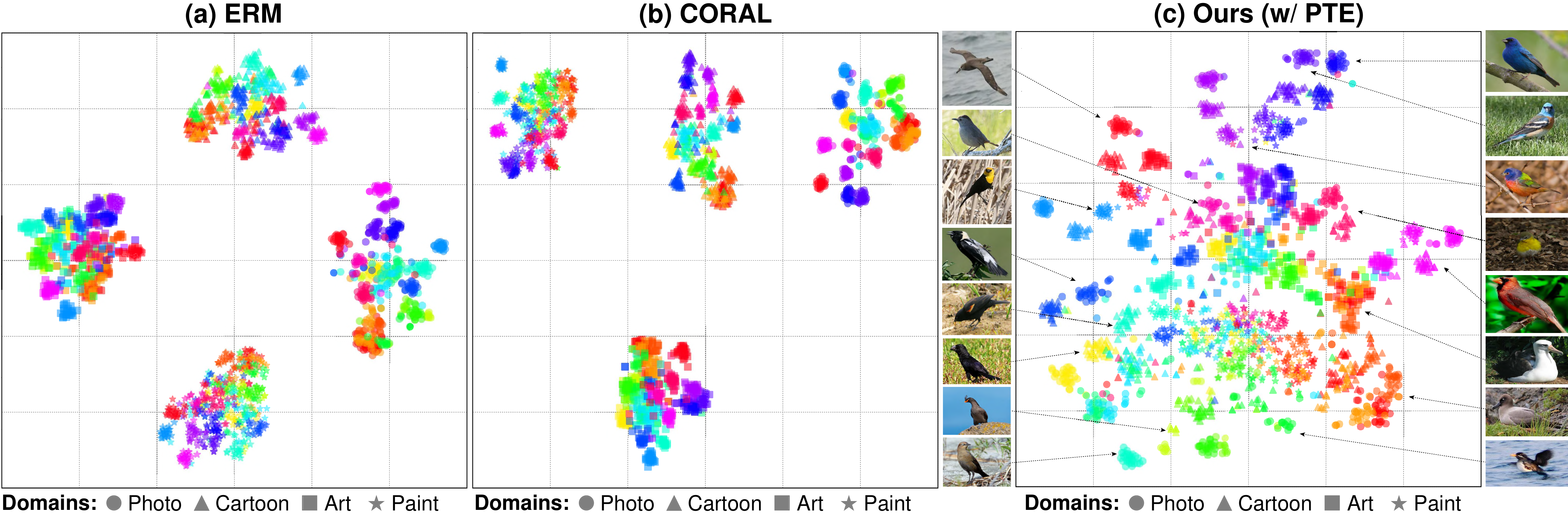}
    \end{center}
    \caption{Visualizations by t-SNE~\cite{van2008visualizing} for (a) ERM~\cite{vapnik1999overview}, (b) CORAL~\cite{sun2016deep}, and (c) ours. We extract latent representations from each model in the multi-source DG setting. We also provide sample images across different classes. Note that we differently color-coded each point according to its class and differently shaped to its domain.}
    \label{fig:tsne}
\end{figure*}

\myparagraph{Qualitative Analysis on the Latent Space}
We use t-SNE~\cite{van2008visualizing} to compute pairwise similarities of embeddings in the latent space and visualize them in a low dimensional space. In Figure~\ref{fig:tsne}, we provide a comparison of t-SNE visualizations of ERM~\cite{vapnik1999overview}, CORAL~\cite{sun2016deep}, and ours. Marker styles and colors indicate the target domain and the ground truth classes, respectively. The more generalizable model should map images belonging to the same class closely even if they are from different domains. We can observe that the baseline models produce scattered multiple clusters for each domain, which confirms that discrepancy between domains is not successfully reduced (see embeddings of the same domain are clustered closely). Ours is not the case for this. Objects from the same class (or similar attributes) but different domains tend to form a merged cluster, making latent representations close to each other in the high-dimensional space. Additionally, in Figure S4, we provide Grad-CAM~\cite{selvaraju2017grad} visualizations which highlight image regions where the model attends to classify the given object.

\begin{figure*}[!t]
    \begin{center}
        \includegraphics[width=\linewidth]{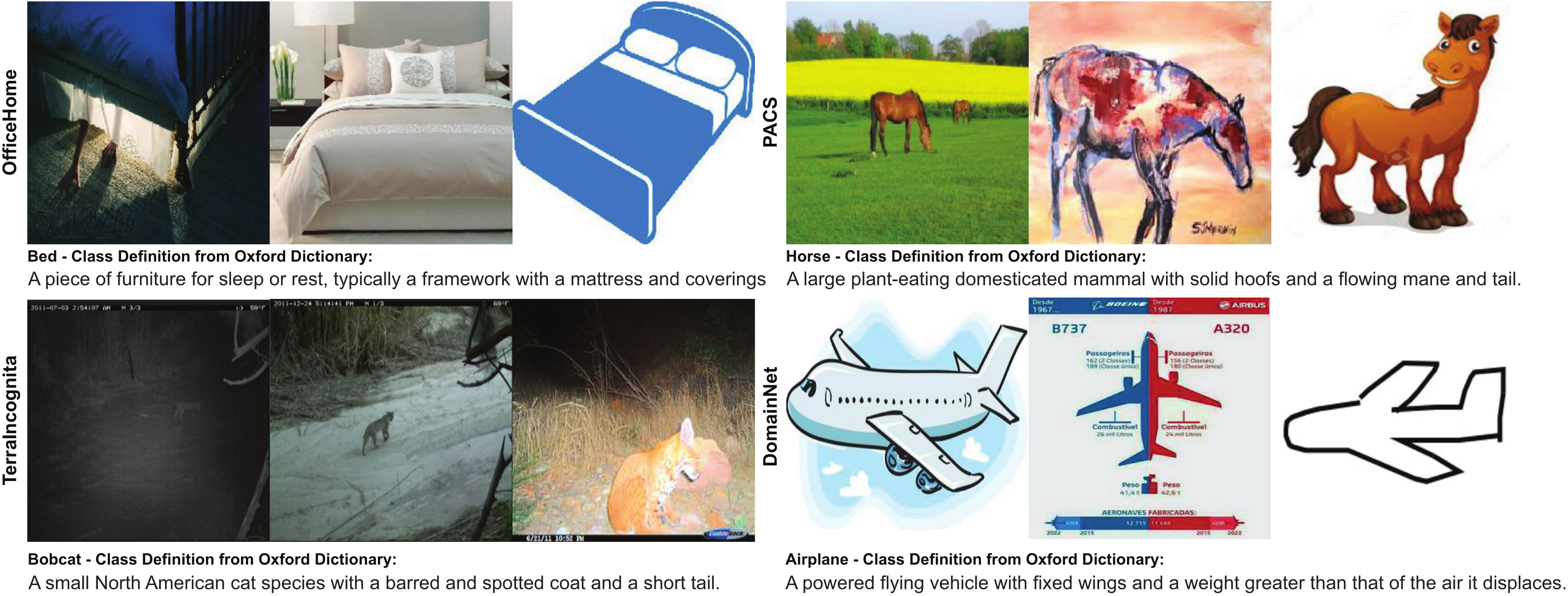}
    \end{center}
    \caption{We use the definition of each class as textual supervision from Oxford English Dictionary~\cite{stevenson2010oxford}. Examples from OfficeHome, PACS, TerraIncognita, and DomainNet on different domains are visualized.}
    \label{fig:domainbed_captions}
\end{figure*}

{
\setlength{\tabcolsep}{4pt}
\renewcommand{\arraystretch}{1.3} 
\begin{table}[t]
	\begin{center}
	    \caption{Average out-of-distribution test accuracies on the DomainBed setting. Here we compare with 14 DG algorithms on the following five multi-domain datastes: VLCS~\cite{fang2013unbiased}, PACS~\cite{Li2017dg}, OfficeHome~\cite{venkateswara2017deep}, TerraIncognita~\cite{beery2018recognition}, and DomainNet~\cite{peng2019moment}. The results of compared DG algorithms are excerpted from DomainBed~\cite{gulrajani2020search}. Note that we use the validation set (from source domains) for the model selection.}
	    \label{Table:domainbed}
    	\resizebox{\linewidth}{!}{%
        \begin{tabular}{lcccccccc}
            \toprule
            \textbf{Algorithm}  &  \textbf{VLCS~\cite{fang2013unbiased}}  &  \textbf{PACS~\cite{Li2017dg}}  &  \textbf{OfficeHome~\cite{venkateswara2017deep}}  &  \textbf{TerraIncognita~\cite{beery2018recognition}}  &  \textbf{DomainNet~\cite{peng2019moment}}  &  \textbf{Avg} \\
            \midrule
            Ours w/ PTE & \textbf{79.0 $\pm$ 0.2} & 85.1 $\pm$ 0.3 & \textbf{70.1 $\pm$ 0.1} & 48.0 $\pm$ 0.2 & \textbf{44.1 $\pm$ 0.1} & \textbf{65.2} \\
            \midrule
            CORAL~\cite{sun2016deep}   &  78.8 $\pm$ 0.6  &  86.2 $\pm$ 0.3  &  68.7 $\pm$ 0.3  &  47.6 $\pm$ 1.0  &  41.5 $\pm$ 0.1  & 64.6 \\
            SagNet~\cite{nam2021reducing}   &  77.8 $\pm$ 0.5  &  \textbf{86.3 $\pm$ 0.2}  &  68.1 $\pm$ 0.1  &  \textbf{48.6 $\pm$ 1.0}  &  40.3 $\pm$ 0.1  & 64.2 \\
            MLDG~\cite{li2018learning}   &  77.2 $\pm$ 0.4  &  84.9 $\pm$ 1.0  &  66.8 $\pm$ 0.6  &  47.7 $\pm$ 0.9  &  41.2 $\pm$ 0.1  & 63.6 \\
            Mixup~\cite{yan2020improve}   &  77.4 $\pm$ 0.6  &  84.6 $\pm$ 0.6  &  68.1 $\pm$ 0.3  &  47.9 $\pm$ 0.8  &  39.2 $\pm$ 0.1  & 63.4 \\
            ERM~\cite{vapnik1999overview}   &  77.5 $\pm$ 0.4  &  85.5 $\pm$ 0.2  &  66.5 $\pm$ 0.3  &  46.1 $\pm$ 1.8  &  40.9 $\pm$ 0.1  & 63.3 \\
            MTL~\cite{blanchard2017domain}   &  77.2 $\pm$ 0.4  &  84.6 $\pm$ 0.5  &  66.4 $\pm$ 0.5  &  45.6 $\pm$ 1.2  &  40.6 $\pm$ 0.1  & 62.9 \\
            RSC~\cite{huangRSC2020}   &  77.1 $\pm$ 0.5  &  85.2 $\pm$ 0.9  &  65.5 $\pm$ 0.9  &  46.6 $\pm$ 1.0  &  38.9 $\pm$ 0.5  & 62.7 \\
            DANN~\cite{ganin2016domain}   &  78.6 $\pm$ 0.4  &  83.6 $\pm$ 0.4  &  65.9 $\pm$ 0.6  &  46.7 $\pm$ 0.5  &  38.3 $\pm$ 0.1  & 62.6 \\
            CDANN~\cite{li2018deep}   &  77.5 $\pm$ 0.1  &  82.6 $\pm$ 0.9  &  65.8 $\pm$ 1.3  &  45.8 $\pm$ 1.6  &  38.3 $\pm$ 0.3  & 62.0 \\
            VREx~\cite{krueger2020out}   &  78.3 $\pm$ 0.2  &  84.9 $\pm$ 0.6  &  66.4 $\pm$ 0.6  &  46.4 $\pm$ 0.6  &  33.6 $\pm$ 2.9  & 61.9 \\
            ARM~\cite{zhang2020adaptive}   &  77.6 $\pm$ 0.3  &  85.1 $\pm$ 0.4  &  64.8 $\pm$ 0.3  &  45.5 $\pm$ 0.3  &  35.5 $\pm$ 0.2  & 61.7 \\
            IRM~\cite{arjovsky2019invariant}   &  78.5 $\pm$ 0.5  &  83.5 $\pm$ 0.8  &  64.3 $\pm$ 2.2  &  47.6 $\pm$ 0.8  &  33.9 $\pm$ 2.8  & 61.6 \\
            GroupDRO~\cite{sagawa2019distributionally}   &  76.7 $\pm$ 0.6  &  84.4 $\pm$ 0.8  &  66.0 $\pm$ 0.7  &  43.2 $\pm$ 1.1  &  33.3 $\pm$ 0.2  & 60.7 \\
            MMD~\cite{li2018domain}   &  77.5 $\pm$ 0.9  &  84.6 $\pm$ 0.5  &  66.3 $\pm$ 0.1  &  42.2 $\pm$ 1.6  &  23.4 $\pm$ 9.5  & 58.8 \\
            \bottomrule
        \end{tabular}}
     \end{center}
\end{table}
}

\myparagraph{Large-Scale Experiments on DomainBed.}
To further verify the effectiveness of the proposed algotithm, we conduct large-scale experiments on DomainBed~\cite{gulrajani2020search}, which is a unified testbed useful for evaluating DG algorithms. We evaluate our algorithm on the following five multi-domain datasets (i.e. VLCS~\cite{fang2013unbiased}, PACS~\cite{Li2017dg}, OfficeHome~\cite{venkateswara2017deep}, TerraIncognita~\cite{beery2018recognition}, and DomainNet~\cite{peng2019moment}) and compare with 14 DG algorithms (i.e. CORAL~\cite{sun2016deep}, SagNet~\cite{nam2021reducing}, MLDG~\cite{li2018learning}, Mixup~\cite{yan2020improve}, ERM~\cite{vapnik1999overview}, MTL~\cite{blanchard2017domain}, RSC~\cite{huangRSC2020}, DANN~\cite{ganin2016domain}, CDANN~\cite{li2018deep}, VREx~\cite{krueger2020out}, ARM~\cite{zhang2020adaptive}, IRM~\cite{arjovsky2019invariant}, GroupDRO~\cite{sagawa2019distributionally}, and MMD~\cite{li2018domain}). 

The results of compared DG algorithms are excerpted from DomainBed~\cite{gulrajani2020search}. For each algorithm, they have conducted a random search of 20 hyperparameter choices. Thus, we also conduct a random search of 20 hyperparameter choices from the following: learning rate from $5\cdot10^{\text{Uniform}(-5, -4)}$, weight decay from $10^{\text{Uniform}(-4, -3)}$, dropout probability from $\text{RandomChoice}([0, 0.1, 0.5])$, and a batch size from $2^{\text{Uniform}(5, 5.5)}$. Other hyperparameters are fixed to the default values. We report averaged results across three independent runs. 

Here, we leverage cross-modality supervision from \textit{Per-Class} texts. Specifically, we use definitions from Oxford English Dictionary~\cite{stevenson2010oxford} to ground visual representations. In Table~\ref{Table:domainbed}, we observe that the proposed algorithm shows state-of-the-art performance, where it ranks 1st in average performance for five multi-domain datasets. Additionally, we provide per-domain results on each dataset in Table S3-S7. We suppose the inferior performance on some datasets is because they often do not contain enough semantic ques that can be aligned with the textual definitions. For example, it is difficult to recognize ``a barred and spotted coat'' from the images of the TerraIncognita dataset in Figure~\ref{fig:domainbed_captions}.

\section{Conclusion}

Towards learning more domain-invariant representations, we advocate for leveraging the cross-modality supervision. Specifically, we propose a new approach where class-discriminative natural language sentence is used during training. {\em Visual and Textual Joint Embedder} encourages learning visual representations that align with the pivot sentence embedding. {\em Textual Explanation Generator} encourages to consistently verbalize why it made a certain prediction with natural language. The experiments with the newly created CUB-DG dataset and the DomainBed benchmarks show that our model outperforms prior work under the standard DG evaluation setting. Our analysis further shows that the text modality can be successfully used to justify visual predictions as well as improve the model's representational generalization power.

\myparagraph{Acknowledgements.}
This work was supported by supported by the National Research Foundation of Korea grant (NRF-2021R1C1C1009608), Basic Science Research Program (NRF-2021R1A6A1A13044830), and ICT Creative Consilience program (IITP-2022-2022-0-01819).

\clearpage
\renewcommand{\thepage}{S\arabic{page}} 
\renewcommand{\thesection}{S\arabic{section}}  
\renewcommand{\thetable}{S\arabic{table}}  
\renewcommand{\thefigure}{S\arabic{figure}}

\title{Supplementary Material for: \\ Grounding Visual Representations with Texts for Domain Generalization}
\titlerunning{Grounding Visual Representations with Texts for Domain Generalization}
\author{
    Seonwoo Min\inst{1} \and
    Nokyung Park \inst{2} \and
    Siwon Kim \inst{3} \and \\
    Seunghyun Park \inst{4} \and
    Jinkyu Kim \inst{2}
}
\authorrunning{Min et al.}
\institute{
    LG AI Research, South Korea \and
    Computer Science and Engineering, Korea University, South Korea \and
    Electrical and Computer Engineering, Seoul National University, South Korea \and
    Clova AI Research, NAVER Corp., South Korea \\
    Correspondence: \email{jinkyukim@korea.ac.kr}
}
\maketitle

This supplementary material contains details of our paper which we could not provide in the main manuscript due to page limits. We provide (1) details of the CUB-DG data split procedure, (2) implementation details, (3) additional single-domain DG results, (4) additional ablation studies results, (5) analysis with Grad-CAM, and (5) detailed DomainBed experiment results.

\begin{figure}[h]
    \begin{center}
        \includegraphics[width=0.8\linewidth]{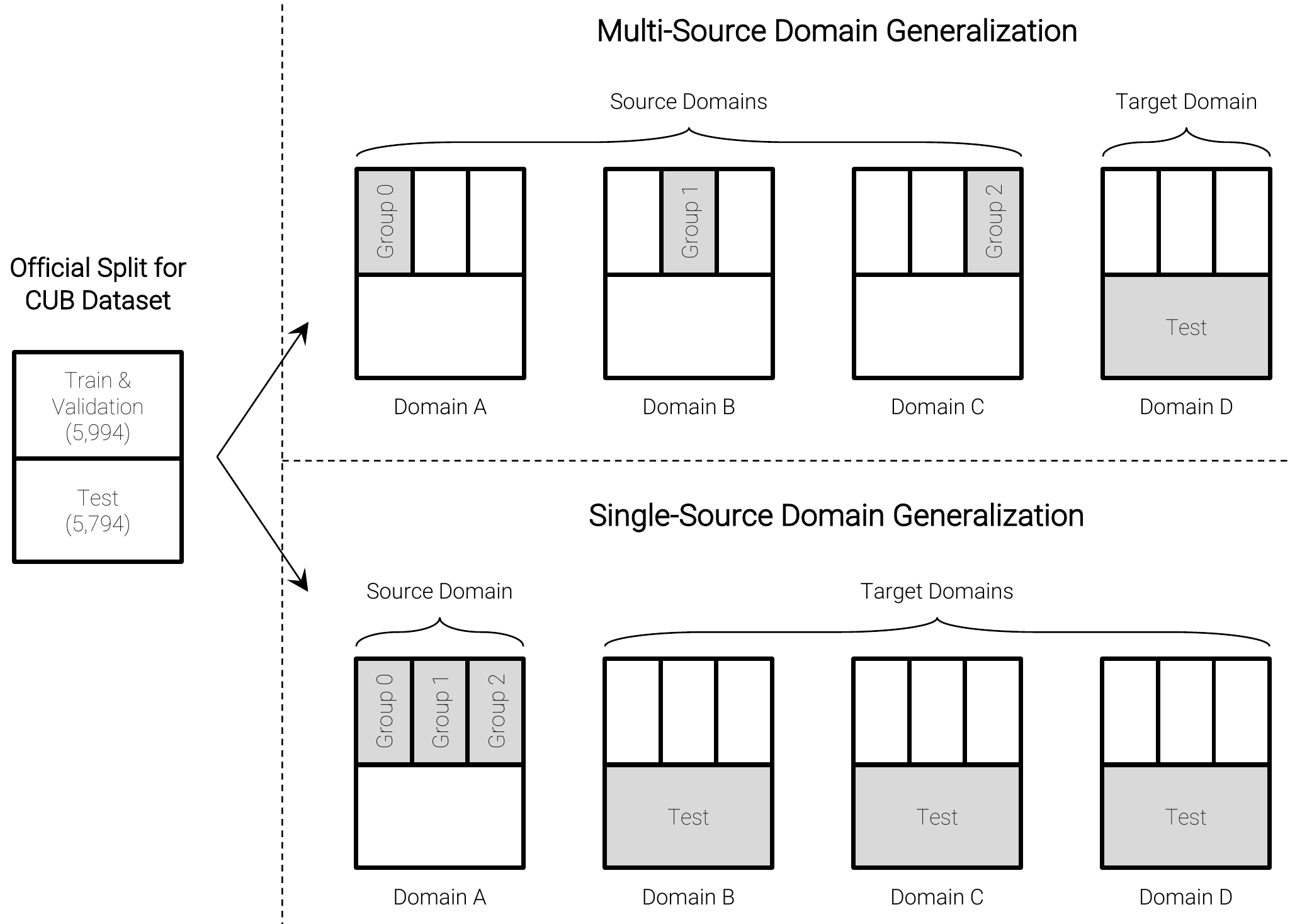}
    \end{center}
    \caption{Data split procedures for the CUB-DG dataset. We start from the official split of the CUB dataset. We divide the train-validation set into three disjoint groups, e.g. Group 0, Group 1, and Group 2. For the multi-source DG task, we select a different group from each source domain (gray boxes), so that the different domains do not share the siblings of the same image. For the single-source DG task, we use all three groups from a source domain.}
    \label{fig:dataset_split}
\end{figure}

\myparagraph{Details of CUB-DG Data Split Procedure.}
In Figure~\ref{fig:dataset_split}, we show an overview of our data split procedure for the CUB-DG dataset. Note that we tried to make it close to real-life scenarios where versions of the same images do not appear in different domains.

\myparagraph{Implementation Details.}
We follow the implementations of DomainBed~\cite{gulrajani2020search}, which is a unified testbed useful for evaluating DG algorithms. We use ResNet-50~\cite{he2016deep} as the backbone of different algorithms. It is pre-trained on ImageNet~\cite{deng2009imagenet} and produces a 2,048-dimensional latent representation from the last layer. We train each DG algorithm for 5,000 steps using Adam optimizer with a batch size of 32 for each source domain. Standard image augmentations (i.e. random cropping, horizontal flipping, color jittering, grayscale conversion, and normalization) are used during the training. For the model and training hyperparameters of each algorithm, we use the default values used in the DomainBed. In our case, we use \num{1.0} for $\lambda_\textnormal{align}$ and \num{1.0} for $\lambda_\textnormal{expl}$. The learning rate is set to $5\mathrm{e}{-5}$ for the backbone parameters and $5\mathrm{e}{-4}$ for the newly introduced parameters.

\begin{figure*}[hb]
    \begin{center}
        \includegraphics[width=\linewidth]{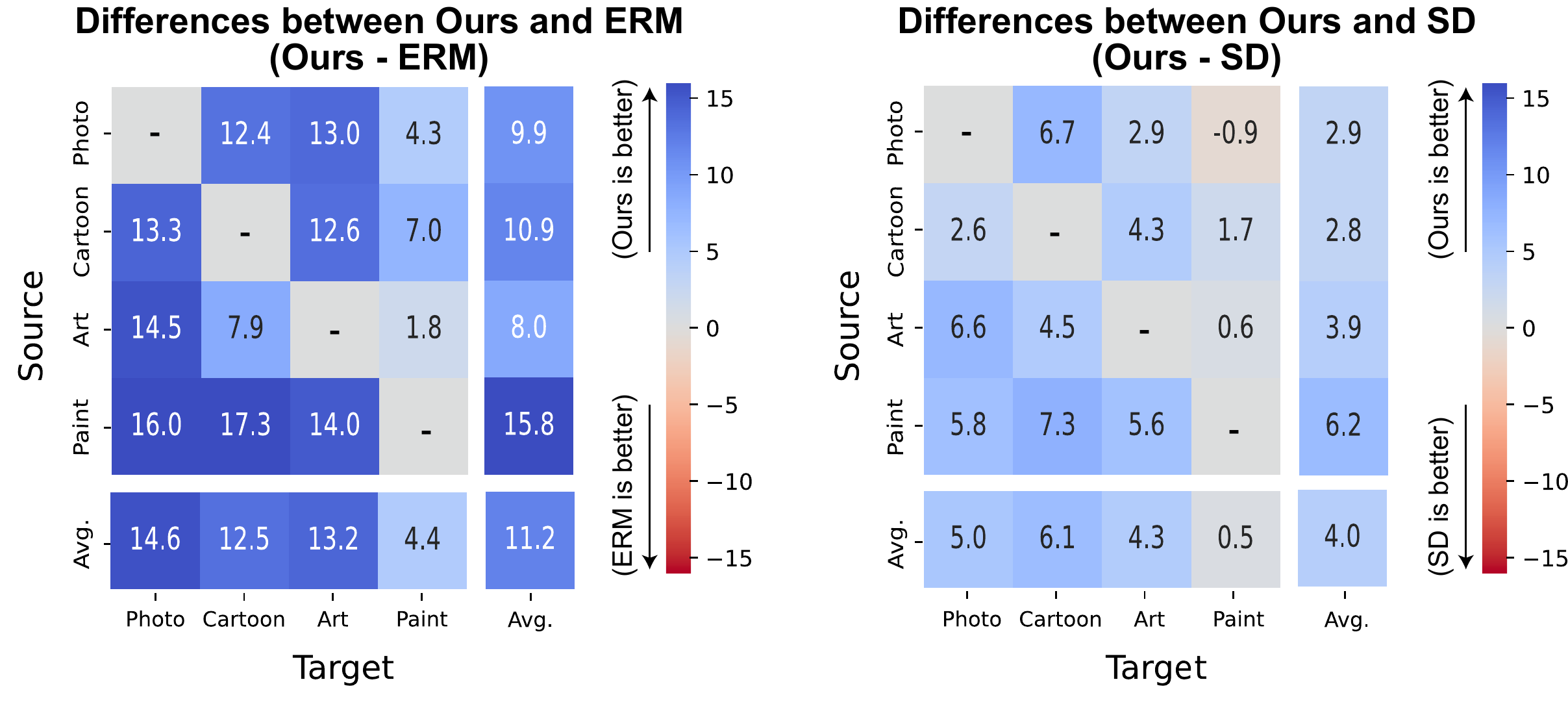}
    \end{center}
    \caption{Out-of-distribution test accuracies in the single-source DG setting where we train our model with a single source domain (rows) and evaluate with other remaining target domains (columns).
    We show performance differences between ours and two baselines (i.e. ERM~\cite{vapnik1999overview} and SD~\cite{pezeshki2020gradient}). {\bf Blue indicates that ours is better.}}
    \label{fig:heatmap}
\end{figure*}

\myparagraph{Additional Single-Source DG Results.}
We provide additional results in the single-source DG task on the CUB-DG dataset. In Figure~\ref{fig:heatmap}, we provide a heatmap that more clearly demonstrates the performance differences between ours and two baselines, i.e., ERM~\cite{vapnik1999overview} and SD~\cite{pezeshki2020gradient}. Each cell contains accuracy differences for source-target combinations, and the color blue indicates that ours performs better. Next, we provide the full results for comparing our model with six DG algorithms. Note that we excluded some algorithms (e.g. CORAL~\cite{sun2016deep} and Mixup~\cite{yan2020improve}). Since those algorithms explicitly match distributions across different domains, they are inapplicable for the single-source DG setting. 

{
\setlength{\tabcolsep}{4pt}
\renewcommand{\arraystretch}{1.3} 
\begin{table}[t]
	\begin{center}
	    \caption{Out-of-distribution test accuracies on the CUB-DG benchmark dataset. Here we compare six DG algorithms in the single-source DG setting. For each target domain, we report the averaged results from three models independently trained with each of the remaining domains. Note that we use the validation set (from source domains) for the model selection.}
	    \label{Table:single-domain-full}
    	\resizebox{.65\linewidth}{!}{%
    	\begin{tabular}{@{}lcccccccccc@{}} 
    	    \toprule
        	\multirow{2}{*}{Model} & \multicolumn{4}{c}{Source Domain} & \multirow{2}{*}{Avg} \\\cmidrule{2-5}
        	& Photo & Cartoon & Art & Paint &  \\ 
        	\midrule
            Ours w/ PTE & \textbf{69.6} & \textbf{48.1} & \textbf{41.1} & 24.0 & \textbf{45.7} \\
            Ours w/ STE & 69.0 & \textbf{48.1} & 39.2 & \textbf{24.9} & 45.3 \\
            \midrule
            SD~\cite{pezeshki2020gradient}   & 64.6 & 41.9 & 36.9 & 23.6 & 41.7 \\
            SagNet~\cite{nam2021reducing}  & 56.0 & 38.1 & 28.7 & 22.2 & 36.3 \\
            VREX~\cite{krueger2020out}  & 55.1 & 36.2 & 27.3 & 19.8 & 34.6 \\
            ERM~\cite{vapnik1999overview}   & 55.0 & 35.6 & 27.9 & 19.7 & 34.5 \\
            ARM~\cite{zhang2020adaptive}   & 54.9 & 36.9 & 28.0 & 20.6 & 35.1 \\
            IRM~\cite{arjovsky2019invariant}   & 53.1 & 35.6 & 27.6 & 19.3 & 33.9 \\
            \bottomrule
        \end{tabular}}
     \end{center}
\end{table}
}

{
\setlength{\tabcolsep}{4pt}
\renewcommand{\arraystretch}{1.3} 
\begin{table}[ht]
	\begin{center}
	    \caption{Results from additional ablation studies. We vary our base model in several directions and measured the performance on the multi-source DG task.}
	    \label{Table:pte_ablation_studies}
    	\resizebox{0.75\linewidth}{!}{
    	\begin{tabular}{@{}lcccccc@{}} 
    	    \toprule
        	& \multirow{2}{*}{\shortstack{Pre-trained\\Textual Encoder}} & \multicolumn{4}{c}{Target Domain} & \multirow{2}{*}{Avg} \\\cmidrule{3-6}
        	& & Photo & Cartoon & Art & Paint &  \\ 
        	\midrule
            Base & CLIP~\cite{radford2021learning} & 74.6 & \textbf{64.2} & \textbf{52.2} & \textbf{37.0} & \textbf{57.0} \\
            \midrule
            \multirow{2}{*}{(C)}  & MPNet~\cite{song2020mpnet} & 74.5 & 63.1 & 49.8 & 37.7 & 56.3 \\
              & DistillBERT~\cite{Sanh2019DistilBERTAD} & 74.2 & 62.2 & 50.4 & 38.4 & 56.3 \\
              & MiniLM~\cite{wang2020minilm} & 73.6 & 64.7 & 51.4 & 35.7 & 56.3 \\
            \bottomrule
        \end{tabular}}
     \end{center}
\end{table}
}

In Table~\ref{Table:single-domain-full}, we report the averaged results from three models independently trained with each of the remaining domains. We observe that the proposed models outperform the others. Cross-modality supervision is especially effective in the single-source DG setting where visual representations alone deliver little information for domain invariances.  

\begin{figure*}[ht]
    \begin{center}
        \includegraphics[width=0.65\linewidth]{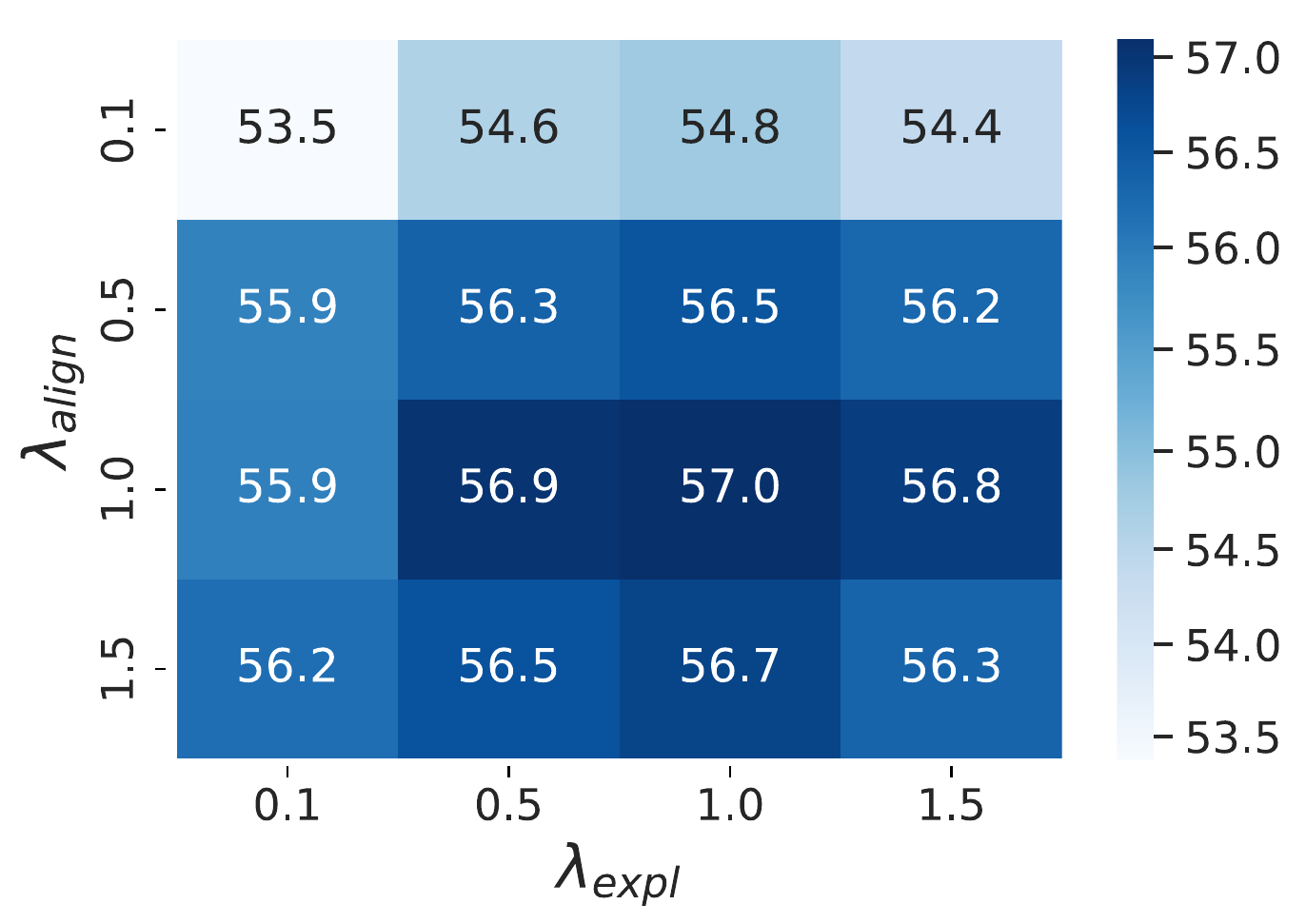}
    \end{center}
    \caption{Additional results from ablation studies. We report out-of-distribution test accuracies in the multi-source DG setting where we train our model with a $\lambda_\textnormal{expl}$ (rows) and $\lambda_\textnormal{align}$ (columns)}
    \label{fig:lambda_ablation}
\end{figure*}

\myparagraph{Additional Ablation Studies Results.}
We provide additional results from the ablation studies. We report averaged results across three independent runs in the multi-source DG setting. In Figure~\ref{fig:lambda_ablation}, we provide a heatmap for more extensive range of $\lambda_\textnormal{expl}$ (rows) and $\lambda_\textnormal{align}$ (columns). Again, we can see that the former is more crucial in the training of our proposed model. In Table~\ref{Table:pte_ablation_studies}, we compare the impact of embeddings from other PTEs, i.e. MPNet~\cite{song2020mpnet}, DistillBERT~\cite{Sanh2019DistilBERTAD}, and MiniLM~\cite{wang2020minilm}. The results show that different PTEs also successfully produce domain-invariant representations.

\begin{figure}[!t]
    \begin{center}
        \includegraphics[width=\linewidth]{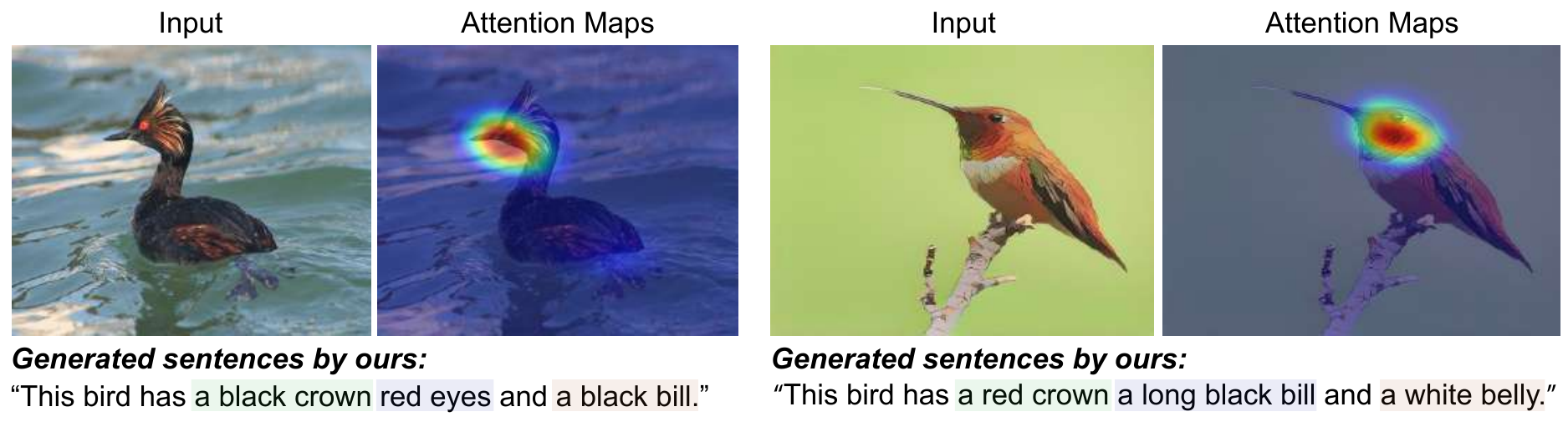}
    \end{center}
    \caption{We provide visualizations of attention maps (i.e. where the model sees) by Grad-CAM for ours as well as the generated sentences.}
    \label{fig:gradcam}
\end{figure}

\myparagraph{Analysis with Grad-CAM.}
As shown in Figure~\ref{fig:gradcam}, we use Grad-CAM~\cite{selvaraju2017grad} to highlight image regions where the model attends to classify the given object. We provide two examples for different target domains (i.e. Cartoon and Photo) where we compare the model's attention maps. We observe that our proposed model captures the class-discriminative features (i.e. short pointy beak), which are compatible with the generated sentence. Note that red is the attended region.

\myparagraph{Detailed DomainBed Experiment Results.}
In Table~\ref{Table:domainbed_vlcs}--\ref{Table:domainbed_domainnet}, we provide per-domain results on each of the five multi-domain datasets from the large-scale DomainBed~\cite{gulrajani2020search} experiments. Following their experiment protocols, we report the averaged results from three independent trials. In each trial, entire random choices (e.g. dataset splits, hyperparameter search, and weight initialization) in the study are renewed. Note that we use the validation set (from source domains) for the model selection.

{
\setlength{\tabcolsep}{4pt}
\renewcommand{\arraystretch}{1.3} 
\begin{table}[t]
	\begin{center}
	    \caption{Per-domain out-of-distribution test accuracies on the VLCS~\cite{fang2013unbiased} dataset. The results of compared DG algorithms are excerpted from DomainBed~\cite{gulrajani2020search}. Note that we use the train domain validation set (from source domains) for the model selection.}
	    \label{Table:domainbed_vlcs}
    	\resizebox{.7\linewidth}{!}{%
        \begin{tabular}{lccccc}
        \toprule
        \textbf{Method}    &  \textbf{Caltech}            &  \textbf{LabelMe}            &  \textbf{SUN09}            &  \textbf{VOC2007}            & \textbf{Avg} \\
        \midrule
        Ours w/ PTE & 98.8 $\pm$ 0.1 & 64.0 $\pm$ 0.3 & \textbf{75.2 $\pm$ 0.5} & \textbf{77.9 $\pm$ 1.0} & \textbf{79.0} \\
        \midrule
        CORAL~\cite{sun2016deep}                 &  98.3 $\pm$ 0.1        &  66.1 $\pm$ 1.2        &  73.4 $\pm$ 0.3        &  77.5 $\pm$ 1.2        & 78.8 \\
        DANN~\cite{ganin2016domain}                  &  \textbf{99.0 $\pm$ 0.3}        &  \textbf{65.1 $\pm$ 1.4}        &  73.1 $\pm$ 0.3        &  77.2 $\pm$ 0.6        & 78.6 \\
        IRM~\cite{arjovsky2019invariant}                   &  98.6 $\pm$ 0.1        &  64.9 $\pm$ 0.9        &  73.4 $\pm$ 0.6        &  77.3 $\pm$ 0.9        & 78.5 \\
        VREx~\cite{krueger2020out}                  &  98.4 $\pm$ 0.3        &  64.4 $\pm$ 1.4        &  74.1 $\pm$ 0.4        &  76.2 $\pm$ 1.3        & 78.3 \\
        SagNet~\cite{nam2021reducing}                &  97.9 $\pm$ 0.4        &  64.5 $\pm$ 0.5        &  71.4 $\pm$ 1.3        &  77.5 $\pm$ 0.5        & 77.8 \\
        ARM~\cite{zhang2020adaptive}                   &  98.7 $\pm$ 0.2        &  63.6 $\pm$ 0.7        &  71.3 $\pm$ 1.2        &  76.7 $\pm$ 0.6        & 77.6 \\
        ERM~\cite{vapnik1999overview}                   &  97.7 $\pm$ 0.4        &  64.3 $\pm$ 0.9        &  73.4 $\pm$ 0.5        &  74.6 $\pm$ 1.3        & 77.5 \\
        MMD~\cite{li2018domain}                   &  97.7 $\pm$ 0.1        &  64.0 $\pm$ 1.1        &  72.8 $\pm$ 0.2        &  75.3 $\pm$ 3.3        & 77.5 \\
        CDANN~\cite{li2018deep}                 &  97.1 $\pm$ 0.3        &  \textbf{65.1 $\pm$ 1.2}        &  70.7 $\pm$ 0.8        &  77.1 $\pm$ 1.5        & 77.5 \\
        Mixup~\cite{yan2020improve}                 &  98.3 $\pm$ 0.6        &  64.8 $\pm$ 1.0        &  72.1 $\pm$ 0.5        &  74.3 $\pm$ 0.8        & 77.4 \\
        MLDG~\cite{li2018learning}                  &  97.4 $\pm$ 0.2        &  65.2 $\pm$ 0.7        &  71.0 $\pm$ 1.4        &  75.3 $\pm$ 1.0        & 77.2 \\
        MTL~\cite{blanchard2017domain}                   &  97.8 $\pm$ 0.4        &  64.3 $\pm$ 0.3        &  71.5 $\pm$ 0.7        &  75.3 $\pm$ 1.7        & 77.2 \\
        RSC~\cite{huangRSC2020}                   &  97.9 $\pm$ 0.1        &  62.5 $\pm$ 0.7        &  72.3 $\pm$ 1.2        &  75.6 $\pm$ 0.8        & 77.1 \\
        GroupDRO~\cite{sagawa2019distributionally}              &  97.3 $\pm$ 0.3        &  63.4 $\pm$ 0.9        &  69.5 $\pm$ 0.8        &  76.7 $\pm$ 0.7        & 76.7 \\
        \bottomrule
        \end{tabular}}
     \end{center}
\end{table}
}

{
\setlength{\tabcolsep}{4pt}
\renewcommand{\arraystretch}{1.3} 
\begin{table}[t]
	\begin{center}
	    \caption{Per-domain out-of-distribution test accuracies on the PACS~\cite{Li2017dg} dataset. The results of other compared DG algorithms are brought from DomainBed~\cite{gulrajani2020search}. Note that we use the train domain validation set (from source domains) for the model selection.}
	    \label{Table:domainbed_pacs}
    	\resizebox{.7\linewidth}{!}{%
        \begin{tabular}{lccccc}
        \toprule
        \textbf{Method} &  \textbf{Art Painting} & \textbf{Cartoon} &  \textbf{Photo} & \textbf{Sketch} & \textbf{Avg} \\
        \midrule
        Ours w/ PTE & 87.9 $\pm$ 0.3 & 78.4 $\pm$ 1.0 & \textbf{98.2 $\pm$ 0.1} & 75.7 $\pm$ 0.4 & 85.1 \\
        \midrule
        SagNet~\cite{nam2021reducing}                &  87.4 $\pm$ 1.0        &  80.7 $\pm$ 0.6        &  97.1 $\pm$ 0.1        &  \textbf{80.0 $\pm$ 0.4}        & \textbf{86.3} \\
        CORAL~\cite{sun2016deep}                 &  \textbf{88.3 $\pm$ 0.2}        &  80.0 $\pm$ 0.5        &  97.5 $\pm$ 0.3        &  78.8 $\pm$ 1.3        & 86.2 \\
        ERM~\cite{vapnik1999overview}                   &  84.7 $\pm$ 0.4        &  \textbf{80.8 $\pm$ 0.6}        &  97.2 $\pm$ 0.3        &  79.3 $\pm$ 1.0        & 85.5 \\
        RSC~\cite{huangRSC2020}                   &  85.4 $\pm$ 0.8        &  79.7 $\pm$ 1.8        &  97.6 $\pm$ 0.3        &  78.2 $\pm$ 1.2        & 85.2 \\
        ARM~\cite{zhang2020adaptive}                   &  86.8 $\pm$ 0.6        &  76.8 $\pm$ 0.5        &  97.4 $\pm$ 0.3        &  79.3 $\pm$ 1.2        & 85.1 \\
        MLDG~\cite{li2018learning}                  &  85.5 $\pm$ 1.4        &  80.1 $\pm$ 1.7        &  97.4 $\pm$ 0.3        &  76.6 $\pm$ 1.1        & 84.9 \\
        VREx~\cite{krueger2020out}                  &  86.0 $\pm$ 1.6        &  79.1 $\pm$ 0.6        &  96.9 $\pm$ 0.5        &  77.7 $\pm$ 1.7        & 84.9 \\
        Mixup~\cite{yan2020improve}                 &  86.1 $\pm$ 0.5        &  78.9 $\pm$ 0.8        &  97.6 $\pm$ 0.1        &  75.8 $\pm$ 1.8        & 84.6 \\
        MMD~\cite{li2018domain}                   &  86.1 $\pm$ 1.4        &  79.4 $\pm$ 0.9        &  96.6 $\pm$ 0.2        &  76.5 $\pm$ 0.5        & 84.6 \\
        MTL~\cite{blanchard2017domain}                   &  87.5 $\pm$ 0.8        &  77.1 $\pm$ 0.5        &  96.4 $\pm$ 0.8        &  77.3 $\pm$ 1.8        & 84.6 \\
        GroupDRO~\cite{sagawa2019distributionally}              &  83.5 $\pm$ 0.9        &  79.1 $\pm$ 0.6        &  96.7 $\pm$ 0.3        &  78.3 $\pm$ 2.0        & 84.4 \\
        DANN~\cite{ganin2016domain}                  &  86.4 $\pm$ 0.8        &  77.4 $\pm$ 0.8        &  97.3 $\pm$ 0.4        &  73.5 $\pm$ 2.3        & 83.6 \\
        IRM~\cite{arjovsky2019invariant}                   &  84.8 $\pm$ 1.3        &  76.4 $\pm$ 1.1        &  96.7 $\pm$ 0.6        &  76.1 $\pm$ 1.0        & 83.5 \\
        CDANN~\cite{li2018deep}                 &  84.6 $\pm$ 1.8        &  75.5 $\pm$ 0.9        &  96.8 $\pm$ 0.3        &  73.5 $\pm$ 0.6        & 82.6 \\
        \bottomrule
        \end{tabular}}
     \end{center}
\end{table}
}

{
\setlength{\tabcolsep}{4pt}
\renewcommand{\arraystretch}{1.3} 
\begin{table}[t]
	\begin{center}
	    \caption{Per-domain out-of-distribution test accuracies on the OfficeHome~\cite{venkateswara2017deep} dataset. The results of other compared DG algorithms are brought from DomainBed~\cite{gulrajani2020search}. Note that we use the train domain validation set (from source domains) for the model selection.}
	    \label{Table:domainbed_office}
    	\resizebox{.7\linewidth}{!}{%
        \begin{tabular}{lccccc}
        \toprule
        \textbf{Method}    &  \textbf{Art}            &  \textbf{Clipart}            &  \textbf{Product}            &  \textbf{Real-world}            & \textbf{Avg} \\
        \midrule
        Ours w/ PTE & \textbf{66.3 $\pm$ 0.1} & \textbf{55.8 $\pm$ 0.4} & \textbf{78.2 $\pm$ 0.4} & \textbf{80.4 $\pm$ 0.2} & \textbf{70.1} \\
        \midrule
        CORAL~\cite{sun2016deep}                 &  65.3 $\pm$ 0.4        &  54.4 $\pm$ 0.5        &  76.5 $\pm$ 0.1        &  78.4 $\pm$ 0.5        & 68.7 \\
        Mixup~\cite{yan2020improve}                 &  62.4 $\pm$ 0.8        &  54.8 $\pm$ 0.6        &  76.9 $\pm$ 0.3        &  78.3 $\pm$ 0.2        & 68.1 \\
        SagNet~\cite{nam2021reducing}                &  63.4 $\pm$ 0.2        &  54.8 $\pm$ 0.4        &  75.8 $\pm$ 0.4        &  78.3 $\pm$ 0.3        & 68.1 \\
        MLDG~\cite{li2018learning}                  &  61.5 $\pm$ 0.9        &  53.2 $\pm$ 0.6        &  75.0 $\pm$ 1.2        &  77.5 $\pm$ 0.4        & 66.8 \\
        ERM~\cite{vapnik1999overview}                   &  61.3 $\pm$ 0.7        &  52.4 $\pm$ 0.3        &  75.8 $\pm$ 0.1        &  76.6 $\pm$ 0.3        & 66.5 \\
        MTL~\cite{blanchard2017domain}                   &  61.5 $\pm$ 0.7        &  52.4 $\pm$ 0.6        &  74.9 $\pm$ 0.4        &  76.8 $\pm$ 0.4        & 66.4 \\
        VREx~\cite{krueger2020out}                  &  60.7 $\pm$ 0.9        &  53.0 $\pm$ 0.9        &  75.3 $\pm$ 0.1        &  76.6 $\pm$ 0.5        & 66.4 \\
        MMD~\cite{li2018domain}                   &  60.4 $\pm$ 0.2        &  53.3 $\pm$ 0.3        &  74.3 $\pm$ 0.1        &  77.4 $\pm$ 0.6        & 66.3 \\
        GroupDRO~\cite{sagawa2019distributionally}              &  60.4 $\pm$ 0.7        &  52.7 $\pm$ 1.0        &  75.0 $\pm$ 0.7        &  76.0 $\pm$ 0.7        & 66.0 \\
        DANN~\cite{ganin2016domain}                  &  59.9 $\pm$ 1.3        &  53.0 $\pm$ 0.3        &  73.6 $\pm$ 0.7        &  76.9 $\pm$ 0.5        & 65.9 \\
        CDANN~\cite{li2018deep}                 &  61.5 $\pm$ 1.4        &  50.4 $\pm$ 2.4        &  74.4 $\pm$ 0.9        &  76.6 $\pm$ 0.8        & 65.8 \\
        RSC~\cite{huangRSC2020}                   &  60.7 $\pm$ 1.4        &  51.4 $\pm$ 0.3        &  74.8 $\pm$ 1.1        &  75.1 $\pm$ 1.3        & 65.5 \\
        ARM~\cite{zhang2020adaptive}                   &  58.9 $\pm$ 0.8        &  51.0 $\pm$ 0.5        &  74.1 $\pm$ 0.1        &  75.2 $\pm$ 0.3        & 64.8 \\
        IRM~\cite{arjovsky2019invariant}                   &  58.9 $\pm$ 2.3        &  52.2 $\pm$ 1.6        &  72.1 $\pm$ 2.9        &  74.0 $\pm$ 2.5        & 64.3 \\
        \bottomrule
        \end{tabular}}
     \end{center}
\end{table}
}

{
\setlength{\tabcolsep}{4pt}
\renewcommand{\arraystretch}{1.3} 
\begin{table}[t]
	\begin{center}
	    \caption{Per-domain out-of-distribution test accuracies on the TerraIncognita~\cite{beery2018recognition} dataset. The results of other compared DG algorithms are brought from DomainBed~\cite{gulrajani2020search}. Note that we use the train domain validation set (from source domains) for the model selection.}
	    \label{Table:domainbed_terra}
    	\resizebox{.7\linewidth}{!}{%
        \begin{tabular}{lccccc}
        \toprule
        \textbf{Method}    &  \textbf{L100}         &  \textbf{L38}          &  \textbf{L43}          &  \textbf{L46}          & \textbf{Avg} \\
        \midrule
        Ours w/ PTE & 53.9 $\pm$ 1.3 & 41.8 $\pm$ 1.2 & \textbf{58.2 $\pm$ 0.9} & 38.0 $\pm$ 0.6 & 48.0 \\
        \midrule
        SagNet~\cite{nam2021reducing}                &  53.0 $\pm$ 2.9        &  43.0 $\pm$ 2.5        &  57.9 $\pm$ 0.6        &  40.4 $\pm$ 1.3        & \textbf{48.6} \\
        Mixup~\cite{yan2020improve}                 &  \textbf{59.6 $\pm$ 2.0}        &  42.2 $\pm$ 1.4        &  55.9 $\pm$ 0.8        &  33.9 $\pm$ 1.4        & 47.9 \\
        MLDG~\cite{li2018learning}                  &  54.2 $\pm$ 3.0        &  \textbf{44.3 $\pm$ 1.1}        &  55.6 $\pm$ 0.3        &  36.9 $\pm$ 2.2        & 47.7 \\
        IRM~\cite{arjovsky2019invariant}                   &  54.6 $\pm$ 1.3        &  39.8 $\pm$ 1.9        &  56.2 $\pm$ 1.8        &  39.6 $\pm$ 0.8        & 47.6 \\
        CORAL~\cite{sun2016deep}                 &  51.6 $\pm$ 2.4        &  42.2 $\pm$ 1.0        &  57.0 $\pm$ 1.0        &  39.8 $\pm$ 2.9        & 47.6 \\
        DANN~\cite{ganin2016domain}                  &  51.1 $\pm$ 3.5        &  40.6 $\pm$ 0.6        &  57.4 $\pm$ 0.5        &  37.7 $\pm$ 1.8        & 46.7 \\
        RSC~\cite{huangRSC2020}                   &  50.2 $\pm$ 2.2        &  39.2 $\pm$ 1.4        &  56.3 $\pm$ 1.4        &  \textbf{40.8 $\pm$ 0.6}        & 46.6 \\
        VREx~\cite{krueger2020out}                  &  48.2 $\pm$ 4.3        &  41.7 $\pm$ 1.3        &  56.8 $\pm$ 0.8        &  38.7 $\pm$ 3.1        & 46.4 \\
        ERM~\cite{vapnik1999overview}                   &  49.8 $\pm$ 4.4        &  42.1 $\pm$ 1.4        &  56.9 $\pm$ 1.8        &  35.7 $\pm$ 3.9        & 46.1 \\
        CDANN~\cite{li2018deep}                 &  47.0 $\pm$ 1.9        &  41.3 $\pm$ 4.8        &  54.9 $\pm$ 1.7        &  39.8 $\pm$ 2.3        & 45.8 \\
        MTL~\cite{blanchard2017domain}                   &  49.3 $\pm$ 1.2        &  39.6 $\pm$ 6.3        &  55.6 $\pm$ 1.1        &  37.8 $\pm$ 0.8        & 45.6 \\
        ARM~\cite{zhang2020adaptive}                   &  49.3 $\pm$ 0.7        &  38.3 $\pm$ 2.4        &  55.8 $\pm$ 0.8        &  38.7 $\pm$ 1.3        & 45.5 \\
        GroupDRO~\cite{sagawa2019distributionally}              &  41.2 $\pm$ 0.7        &  38.6 $\pm$ 2.1        &  56.7 $\pm$ 0.9        &  36.4 $\pm$ 2.1        & 43.2 \\
        MMD~\cite{li2018domain}                   &  41.9 $\pm$ 3.0        &  34.8 $\pm$ 1.0        &  57.0 $\pm$ 1.9        &  35.2 $\pm$ 1.8        & 42.2 \\
        \bottomrule
        \end{tabular}}
     \end{center}
\end{table}
}

{
\setlength{\tabcolsep}{4pt}
\renewcommand{\arraystretch}{1.3} 
\begin{table}[t]
	\begin{center}
	    \caption{Per-domain out-of-distribution test accuracies on the DomainNet~\cite{peng2019moment} dataset. The results of other compared DG algorithms are brought from DomainBed~\cite{gulrajani2020search}. Note that we use the train domain validation set (from source domains) for the model selection.}
	    \label{Table:domainbed_domainnet}
    	\resizebox{.9\linewidth}{!}{%
        \begin{tabular}{lccccccc}
        \toprule
        \textbf{Method}    &  \textbf{Clip}         &  \textbf{Info}         &  \textbf{Paint}        &  \textbf{Quick}        &  \textbf{Real}         &  \textbf{Sketch}       & \textbf{Avg} \\
        \midrule
        Ours w/ PTE & \textbf{62.4 $\pm$ 0.4} & \textbf{21.0 $\pm$ 0.0} & \textbf{50.5 $\pm$ 0.4} & \textbf{13.8 $\pm$ 0.3} & \textbf{64.6 $\pm$ 0.4} & \textbf{52.4 $\pm$ 0.2} & \textbf{44.1} \\
        \midrule
        CORAL~\cite{sun2016deep}                 &  59.2 $\pm$ 0.1        &  19.7 $\pm$ 0.2        &  46.6 $\pm$ 0.3        &  13.4 $\pm$ 0.4        &  59.8 $\pm$ 0.2        &  50.1 $\pm$ 0.6        & 41.5 \\
        MLDG~\cite{li2018learning}                  &  59.1 $\pm$ 0.2        &  19.1 $\pm$ 0.3        &  45.8 $\pm$ 0.7        &  13.4 $\pm$ 0.3        &  59.6 $\pm$ 0.2        &  50.2 $\pm$ 0.4        & 41.2 \\
        ERM~\cite{vapnik1999overview}                   &  58.1 $\pm$ 0.3        &  18.8 $\pm$ 0.3        &  46.7 $\pm$ 0.3        &  12.2 $\pm$ 0.4        &  59.6 $\pm$ 0.1        &  49.8 $\pm$ 0.4        & 40.9 \\
        MTL~\cite{blanchard2017domain}                   &  57.9 $\pm$ 0.5        &  18.5 $\pm$ 0.4        &  46.0 $\pm$ 0.1        &  12.5 $\pm$ 0.1        &  59.5 $\pm$ 0.3        &  49.2 $\pm$ 0.1        & 40.6 \\
        SagNet~\cite{nam2021reducing}                &  57.7 $\pm$ 0.3        &  19.0 $\pm$ 0.2        &  45.3 $\pm$ 0.3        &  12.7 $\pm$ 0.5        &  58.1 $\pm$ 0.5        &  48.8 $\pm$ 0.2        & 40.3 \\
        Mixup~\cite{yan2020improve}                 &  55.7 $\pm$ 0.3        &  18.5 $\pm$ 0.5        &  44.3 $\pm$ 0.5        &  12.5 $\pm$ 0.4        &  55.8 $\pm$ 0.3        &  48.2 $\pm$ 0.5        & 39.2 \\
        RSC~\cite{huangRSC2020}                   &  55.0 $\pm$ 1.2        &  18.3 $\pm$ 0.5        &  44.4 $\pm$ 0.6        &  12.2 $\pm$ 0.2        &  55.7 $\pm$ 0.7        &  47.8 $\pm$ 0.9        & 38.9 \\
        DANN~\cite{ganin2016domain}                  &  53.1 $\pm$ 0.2        &  18.3 $\pm$ 0.1        &  44.2 $\pm$ 0.7        &  11.8 $\pm$ 0.1        &  55.5 $\pm$ 0.4        &  46.8 $\pm$ 0.6        & 38.3 \\
        CDANN~\cite{li2018deep}                 &  54.6 $\pm$ 0.4        &  17.3 $\pm$ 0.1        &  43.7 $\pm$ 0.9        &  12.1 $\pm$ 0.7        &  56.2 $\pm$ 0.4        &  45.9 $\pm$ 0.5        & 38.3 \\
        ARM~\cite{zhang2020adaptive}                   &  49.7 $\pm$ 0.3        &  16.3 $\pm$ 0.5        &  40.9 $\pm$ 1.1        &  9.4 $\pm$ 0.1         &  53.4 $\pm$ 0.4        &  43.5 $\pm$ 0.4        & 35.5 \\
        IRM~\cite{arjovsky2019invariant}                   &  48.5 $\pm$ 2.8        &  15.0 $\pm$ 1.5        &  38.3 $\pm$ 4.3        &  10.9 $\pm$ 0.5        &  48.2 $\pm$ 5.2        &  42.3 $\pm$ 3.1        & 33.9 \\
        VREx~\cite{krueger2020out}                  &  47.3 $\pm$ 3.5        &  16.0 $\pm$ 1.5        &  35.8 $\pm$ 4.6        &  10.9 $\pm$ 0.3        &  49.6 $\pm$ 4.9        &  42.0 $\pm$ 3.0        & 33.6 \\
        GroupDRO~\cite{sagawa2019distributionally}              &  47.2 $\pm$ 0.5        &  17.5 $\pm$ 0.4        &  33.8 $\pm$ 0.5        &  9.3 $\pm$ 0.3         &  51.6 $\pm$ 0.4        &  40.1 $\pm$ 0.6        & 33.3 \\
        MMD~\cite{li2018domain}                   &  32.1 $\pm$ 13.3       &  11.0 $\pm$ 4.6        &  26.8 $\pm$ 11.3       &  8.7 $\pm$ 2.1         &  32.7 $\pm$ 13.8       &  28.9 $\pm$ 11.9       & 23.4 \\
        \bottomrule
        \end{tabular}}
     \end{center}
\end{table}
}

\clearpage

\bibliographystyle{splncs04}
\bibliography{egbib}

\begin{thebibliography}{10}
\providecommand{\url}[1]{\texttt{#1}}
\providecommand{\urlprefix}{URL }
\providecommand{\doi}[1]{https://doi.org/#1}

\bibitem{arjovsky2019invariant}
Arjovsky, M., Bottou, L., Gulrajani, I., Lopez-Paz, D.: Invariant risk
  minimization. arXiv preprint arXiv:1907.02893  (2019)

\bibitem{beery2018recognition}
Beery, S., Van~Horn, G., Perona, P.: Recognition in terra incognita. In:
  Proceedings of the European conference on computer vision (ECCV). pp.
  456--473 (2018)

\bibitem{blanchard2017domain}
Blanchard, G., Deshmukh, A.A., Dogan, U., Lee, G., Scott, C.: Domain
  generalization by marginal transfer learning. arXiv preprint arXiv:1711.07910
   (2017)

\bibitem{cha2021domain}
Cha, J., Cho, H., Lee, K., Park, S., Lee, Y., Park, S.: Domain generalization
  needs stochastic weight averaging for robustness on domain shifts. arXiv
  preprint arXiv:2102.08604  (2021)

\bibitem{deng2009imagenet}
Deng, J., Dong, W., Socher, R., Li, L.J., Li, K., Fei-Fei, L.: Imagenet: A
  large-scale hierarchical image database. In: 2009 IEEE conference on computer
  vision and pattern recognition. pp. 248--255. Ieee (2009)

\bibitem{dosovitskiy2020image}
Dosovitskiy, A., Beyer, L., Kolesnikov, A., Weissenborn, D., Zhai, X.,
  Unterthiner, T., Dehghani, M., Minderer, M., Heigold, G., Gelly, S., et~al.:
  An image is worth 16x16 words: Transformers for image recognition at scale.
  arXiv preprint arXiv:2010.11929  (2020)

\bibitem{fang2013unbiased}
Fang, C., Xu, Y., Rockmore, D.N.: Unbiased metric learning: On the utilization
  of multiple datasets and web images for softening bias. In: Proceedings of
  the IEEE International Conference on Computer Vision (ICCV). pp. 1657--1664
  (2013)

\bibitem{ganea2008transfer}
Ganea, P.A., Pickard, M.B., DeLoache, J.S.: Transfer between picture books and
  the real world by very young children. Journal of cognition and development
  \textbf{9}(1),  46--66 (2008)

\bibitem{ganin2016domain}
Ganin, Y., Ustinova, E., Ajakan, H., Germain, P., Larochelle, H., Laviolette,
  F., Marchand, M., Lempitsky, V.: Domain-adversarial training of neural
  networks. The journal of machine learning research  \textbf{17}(1),
  2096--2030 (2016)

\bibitem{geirhos2018imagenettrained}
Geirhos, R., et~al.: {ImageNet}-trained {CNN}s are biased towards texture. In:
  ICLR (2019), \url{https://openreview.net/forum?id=Bygh9j09KX}

\bibitem{gulrajani2020search}
Gulrajani, I., Lopez-Paz, D.: In search of lost domain generalization. arXiv
  preprint arXiv:2007.01434  (2020)

\bibitem{gunning2017explainable}
Gunning, D.: Explainable artificial intelligence (xai). Defense Advanced
  Research Projects Agency (DARPA)  (2017)

\bibitem{he2016deep}
He, K., Zhang, X., Ren, S., Sun, J.: Deep residual learning for image
  recognition. In: Proceedings of the IEEE Conference on Computer Vision and
  Pattern Recognition (CVPR). pp. 770--778 (2016)

\bibitem{hendricks2016generating}
Hendricks, L.A., Akata, Z., Rohrbach, M., Donahue, J., Schiele, B., Darrell,
  T.: Generating visual explanations. In: European conference on computer
  vision. pp. 3--19. Springer (2016)

\bibitem{hendricks2018grounding}
Hendricks, L.A., Hu, R., Darrell, T., Akata, Z.: Grounding visual explanations.
  In: ECCV (2018)

\bibitem{huangRSC2020}
Huang, Z., Wang, H., Xing, E.P., Huang, D.: Self-challenging improves
  cross-domain generalization. In: Proceedings of the European Conference on
  Computer Vision (ECCV) (2020)

\bibitem{kim2021selfreg}
Kim, D., Yoo, Y., Park, S., Kim, J., Lee, J.: Selfreg: Self-supervised
  contrastive regularization for domain generalization. In: Proceedings of the
  IEEE/CVF International Conference on Computer Vision. pp. 9619--9628 (2021)

\bibitem{kim2020interpretation}
Kim, S., Yi, J., Kim, E., Yoon, S.: Interpretation of nlp models through input
  marginalization. arXiv preprint arXiv:2010.13984  (2020)

\bibitem{krueger2020out}
Krueger, D., Caballero, E., Jacobsen, J.H., Zhang, A., Binas, J., Zhang, D.,
  Priol, R.L., Courville, A.: Out-of-distribution generalization via risk
  extrapolation (rex). arXiv preprint arXiv:2003.00688  (2020)

\bibitem{lavie2005meteor}
Lavie, A., Agarwal, A.: Meteor: An automatic metric for mt evaluation with
  improved correlation with human judgments. In: EMNLP (2005)

\bibitem{Li2017dg}
Li, D., Yang, Y., Song, Y.Z., Hospedales, T.: Deeper, broader and artier domain
  generalization. In: Proceedings of the IEEE International Conference on
  Computer Vision (ICCV) (2017)

\bibitem{li2018learning}
Li, D., Yang, Y., Song, Y.Z., Hospedales, T.: Learning to generalize:
  Meta-learning for domain generalization. In: Proceedings of the AAAI
  Conference on Artificial Intelligence. vol.~32 (2018)

\bibitem{li2018domain}
Li, H., Pan, S.J., Wang, S., Kot, A.C.: Domain generalization with adversarial
  feature learning. In: Proceedings of the IEEE Conference on Computer Vision
  and Pattern Recognition (CVPR). pp. 5400--5409 (2018)

\bibitem{li2018deep}
Li, Y., Tian, X., Gong, M., Liu, Y., Liu, T., Zhang, K., Tao, D.: Deep domain
  generalization via conditional invariant adversarial networks. In:
  Proceedings of the European Conference on Computer Vision (ECCV). pp.
  624--639 (2018)

\bibitem{lin2004rouge}
Lin, C.Y.: Rouge: A package for automatic evaluation of summaries. In: Text
  summarization branches out. pp. 74--81 (2004)

\bibitem{van2008visualizing}
Van~der Maaten, L., Hinton, G.: Visualizing data using t-sne. Journal of
  machine learning research  \textbf{9}(11) (2008)

\bibitem{muandet2013domain}
Muandet, K., Balduzzi, D., Sch{\"o}lkopf, B.: Domain generalization via
  invariant feature representation. In: Proceedings of the International
  Conference on Machine Learning (ICML). pp. 10--18. PMLR (2013)

\bibitem{nam2021reducing}
Nam, H., et~al.: Reducing domain gap by reducing style bias. In: CVPR (2021)

\bibitem{papineni2002bleu}
Papineni, K., Roukos, S., Ward, T., Zhu, W.J.: Bleu: a method for automatic
  evaluation of machine translation. In: ACL (2002)

\bibitem{peng2019moment}
Peng, X., Bai, Q., Xia, X., Huang, Z., Saenko, K., Wang, B.: Moment matching
  for multi-source domain adaptation. In: Proceedings of the IEEE International
  Conference on Computer Vision (ICCV). pp. 1406--1415 (2019)

\bibitem{pezeshki2020gradient}
Pezeshki, M., Kaba, S.O., Bengio, Y., Courville, A., Precup, D., Lajoie, G.:
  Gradient starvation: A learning proclivity in neural networks. arXiv preprint
  arXiv:2011.09468  (2020)

\bibitem{radford2021learning}
Radford, A., Kim, J.W., Hallacy, C., Ramesh, A., Goh, G., Agarwal, S., Sastry,
  G., Askell, A., Mishkin, P., Clark, J., et~al.: Learning transferable visual
  models from natural language supervision. arXiv preprint arXiv:2103.00020
  (2021)

\bibitem{reed2016learning}
Reed, S., Akata, Z., Lee, H., Schiele, B.: Learning deep representations of
  fine-grained visual descriptions. In: Proceedings of the IEEE conference on
  computer vision and pattern recognition. pp. 49--58 (2016)

\bibitem{ruan2022optimal}
Ruan, Y., Dubois, Y., Maddison, C.J.: Optimal representations for covariate
  shift. In: International Conference on Learning Representations (2022),
  \url{https://openreview.net/forum?id=Rf58LPCwJj0}

\bibitem{sagawa2019distributionally}
Sagawa, S., Koh, P.W., Hashimoto, T.B., Liang, P.: Distributionally robust
  neural networks for group shifts: On the importance of regularization for
  worst-case generalization. arXiv preprint arXiv:1911.08731  (2019)

\bibitem{Sanh2019DistilBERTAD}
Sanh, V., Debut, L., Chaumond, J., Wolf, T.: Distilbert, a distilled version of
  bert: smaller, faster, cheaper and lighter. ArXiv  \textbf{abs/1910.01108}
  (2019)

\bibitem{selvaraju2017grad}
Selvaraju, R.R., Cogswell, M., Das, A., Vedantam, R., Parikh, D., Batra, D.:
  Grad-cam: Visual explanations from deep networks via gradient-based
  localization. In: ICCV. pp. 618--626 (2017)

\bibitem{sinha2017certifying}
Sinha, A., Namkoong, H., Volpi, R., Duchi, J.: Certifying some distributional
  robustness with principled adversarial training. ICLR  (2017)

\bibitem{song2020mpnet}
Song, K., Tan, X., Qin, T., Lu, J., Liu, T.Y.: Mpnet: Masked and permuted
  pre-training for language understanding. Advances in Neural Information
  Processing Systems  \textbf{33},  16857--16867 (2020)

\bibitem{stevenson2010oxford}
Stevenson, A.: Oxford dictionary of English. Oxford University Press, USA
  (2010)

\bibitem{sun2016deep}
Sun, B., Saenko, K.: Deep coral: Correlation alignment for deep domain
  adaptation. In: Proceedings of the European Conference on Computer Vision
  (ECCV). pp. 443--450. Springer (2016)

\bibitem{vapnik1998statistical}
Vapnik, V.: Statistical learning theory new york. NY: Wiley  (1998)

\bibitem{vapnik1999overview}
Vapnik, V.N.: An overview of statistical learning theory. IEEE transactions on
  neural networks  \textbf{10}(5),  988--999 (1999)

\bibitem{vaswani2017attention}
Vaswani, A., Shazeer, N., Parmar, N., Uszkoreit, J., Jones, L., Gomez, A.N.,
  Kaiser, {\L}., Polosukhin, I.: Attention is all you need. In: Advances in
  neural information processing systems. pp. 5998--6008 (2017)

\bibitem{vedantam2015cider}
Vedantam, R., Lawrence~Zitnick, C., Parikh, D.: Cider: Consensus-based image
  description evaluation. In: ICCV (2015)

\bibitem{venkateswara2017deep}
Venkateswara, H., Eusebio, J., Chakraborty, S., Panchanathan, S.: Deep hashing
  network for unsupervised domain adaptation. In: Proceedings of the IEEE
  conference on computer vision and pattern recognition. pp. 5018--5027 (2017)

\bibitem{wang2018deep}
Wang, D., Devin, C., Cai, Q.Z., Yu, F., Darrell, T.: Deep object centric
  policies for autonomous driving. ICRA  (2019)

\bibitem{wang2020minilm}
Wang, W., Wei, F., Dong, L., Bao, H., Yang, N., Zhou, M.: Minilm: Deep
  self-attention distillation for task-agnostic compression of pre-trained
  transformers (2020)

\bibitem{wang2020learning}
Wang, X., Yu, J.: Learning to cartoonize using white-box cartoon
  representations. In: Proceedings of the IEEE/CVF Conference on Computer
  Vision and Pattern Recognition. pp. 8090--8099 (2020)

\bibitem{wang2020heterogeneous}
Wang, Y., Li, H., Kot, A.C.: Heterogeneous domain generalization via domain
  mixup. In: ICASSP 2020-2020 IEEE International Conference on Acoustics,
  Speech and Signal Processing (ICASSP). pp. 3622--3626. IEEE (2020)

\bibitem{WelinderEtal2010}
Welinder, P., Branson, S., Mita, T., Wah, C., Schroff, F., Belongie, S.,
  Perona, P.: {Caltech-UCSD Birds 200}. Tech. Rep. CNS-TR-2010-001, California
  Institute of Technology (2010)

\bibitem{williams1992simple}
Williams, R.J.: Simple statistical gradient-following algorithms for
  connectionist reinforcement learning. Machine learning  \textbf{8}(3),
  229--256 (1992)

\bibitem{wu2018faithful}
Wu, J., Mooney, R.J.: Faithful multimodal explanation for visual question
  answering. arXiv preprint arXiv:1809.02805  (2018)

\bibitem{xu2020adversarial}
Xu, M., Zhang, J., Ni, B., Li, T., Wang, C., Tian, Q., Zhang, W.: Adversarial
  domain adaptation with domain mixup. In: Proceedings of the AAAI Conference
  on Artificial Intelligence. vol.~34, pp. 6502--6509 (2020)

\bibitem{yan2020improve}
Yan, S., Song, H., Li, N., Zou, L., Ren, L.: Improve unsupervised domain
  adaptation with mixup training. arXiv preprint arXiv:2001.00677  (2020)

\bibitem{zeiler2014visualizing}
Zeiler, M.D., Fergus, R.: Visualizing and understanding convolutional networks.
  In: European Conference on Computer Vision. pp. 818--833. Springer (2014)

\bibitem{zhang2020adaptive}
Zhang, M., Marklund, H., Gupta, A., Levine, S., Finn, C.: Adaptive risk
  minimization: A meta-learning approach for tackling group shift. arXiv
  preprint arXiv:2007.02931  (2020)

\bibitem{zhou2016learning}
Zhou, B., Khosla, A., Lapedriza, A., Oliva, A., Torralba, A.: Learning deep
  features for discriminative localization. In: CVPR. pp. 2921--2929 (2016)

\bibitem{zhou2020domain}
Zhou, K., Yang, Y., Qiao, Y., Xiang, T.: Domain generalization with mixstyle.
  In: International Conference on Learning Representations (2020)

\bibitem{zhu2017unpaired}
Zhu, J.Y., Park, T., Isola, P., Efros, A.A.: Unpaired image-to-image
  translation using cycle-consistent adversarial networks. In: Proceedings of
  the IEEE international conference on computer vision. pp. 2223--2232 (2017)

\bibitem{zou2021stylized}
Zou, Z., Shi, T., Qiu, S., Yuan, Y., Shi, Z.: Stylized neural painting. In:
  Proceedings of the IEEE/CVF Conference on Computer Vision and Pattern
  Recognition. pp. 15689--15698 (2021)

\end{thebibliography}

\end{document}